\documentclass[runningheads]{llncs}

\usepackage{eccv}

\usepackage{eccvabbrv}

\usepackage{graphicx}
\usepackage{booktabs}

\usepackage[hyphens]{url}
\usepackage{graphicx}
\usepackage{caption}
\usepackage{makecell}
\usepackage{multirow}
\usepackage{booktabs}
\usepackage{subcaption}
\usepackage[table]{xcolor}
\usepackage{amsfonts}
\usepackage{nicefrac}
\usepackage{microtype}
\usepackage{xcolor}  
\usepackage{colortbl}
\usepackage{amsmath} 
\usepackage{caption}
\usepackage{pifont}
\usepackage{arydshln}
\usepackage{fontawesome}
\usepackage{svg}
\usepackage{graphicx}
\usepackage{wrapfig}
\usepackage{float}
\usepackage{tcolorbox}
\usepackage{adjustbox}
\usepackage{enumitem}
\usepackage[flushleft]{threeparttable}
\usepackage{graphicx}
\usepackage{xspace}
\usepackage{xcolor}  
\usepackage{placeins}

\usepackage[accsupp]{axessibility}  % Improves PDF readability for those with disabilities.

\usepackage{hyperref}

\usepackage{orcidlink}

\begin{document}

% ---------------------------------------------------------------
% TODO REVIEW: Replace with your title
\title{Magic-MM-Embedding: Towards Visual-Token-Efficient Universal Multimodal Embedding with MLLMs} 

% TODO REVIEW: If the paper title is too long for the running head, you can set
% an abbreviated paper title here. If not, comment out.
\titlerunning{Magic-MM-Embedding}

% TODO FINAL: Replace with your author list. 
% Include the authors' OCRID for the camera-ready version, if at all possible.
\author{Qi Li \and
Yanzhe Zhao \and
Yongxin Zhou \and
Yameng Wang \and
Yandong Yang \and \\
Yuanjia Zhou \and
Jinxiang Liu\thanks{Corresponding author.}}

% TODO FINAL: Replace with an abbreviated list of authors.
\authorrunning{Q.~Li et al.}
% First names are abbreviated in the running head.
% If there are more than two authors, 'et al.' is used.

% TODO FINAL: Replace with your institution list.
\institute{Honor Device Co., Ltd., China\\
\email{\{liqi20, zhaoyanzhe2, zhouyongxin, wangyameng, yangyandong,\\
zhouyuanjia2, liujinxiang\}@honor.com}
}

\maketitle

\begin{abstract}
Multimodal Large Language Models (MLLMs) have shown immense promise in universal multimodal retrieval, which aims to find relevant items of various modalities for a given query. However, their practical application is often hindered by the substantial computational cost incurred from processing a large number of tokens from visual inputs. In this paper, we propose Magic-MM-Embedding, a series of novel models that achieve both high efficiency and state-of-the-art performance in universal multimodal embedding. Our approach is built on two synergistic pillars: (1) a highly efficient MLLM architecture incorporating visual token compression to drastically reduce inference latency and training time, and (2) a multi-stage progressive training strategy designed to not only recover but significantly boost performance. This coarse-to-fine training paradigm begins with extensive continued training to restore multimodal understanding and generation capabilities, progresses to large-scale contrastive pretraining and hard negative mining to enhance discriminative power, and culminates in a task-aware fine-tuning stage guided by an MLLM-as-a-Judge for precise data curation. Comprehensive experiments show that our model outperforms existing methods by a large margin while being more inference-efficient.

\end{abstract}    
\section{Introduction}

Multimodal embedding models are designed to project heterogeneous data modalities, such as text, images, and interleaved image-text data, into a unified semantic space. These models are widely applied across various domains, such as multimodal search~\cite{wei2024uniir,magiclens,faysse2024colpali,ma2024unifying}, recommendation systems~\cite{zhang2025notellm,giahi2025vl}, and retrieval-augmented generation~\cite{VisRAG,jeong2025videorag,ma2025visa}. Recently, the field has witnessed a significant paradigm shift, moving beyond the dual-tower architectures such as CLIP~\cite{clip2021Radford} and UniIR~\cite{wei2024uniir}, towards Multimodal Large Language Models (MLLMs)~\cite{bai2025qwen3,li2024llavaone} with stronger multimodal understanding capabilities.

This transition is driven by the intrinsic limitations of dual-tower frameworks: (1) modal-independent encoding architectures with feature post-fusion, which lack deep cross-modal interaction, limit their ability to perform fine-grained multimodal reasoning~\cite{VLM2Vec,E5V,gu2025break}. (2) limited language understanding ability, in which rigid context constraints and limited prior knowledge restrict the understanding of complex semantics~\cite{longclip,flame,yuksekgonul2022and}. In contrast, MLLM-based methods~\cite{E5V,VLM2Vec} treat visual features as discrete tokens, processed jointly with text in a unified transformer. This facilitates deep token-level cross-modal fusion, rather than shallow global alignment. Leveraging this design, along with the extensive world knowledge and robust instruction-following capabilities of LLMs, these models can perform complex multimodal retrieval in diverse scenarios.

\begin{figure}[tb]
  \centering
  \includegraphics[width=\linewidth]{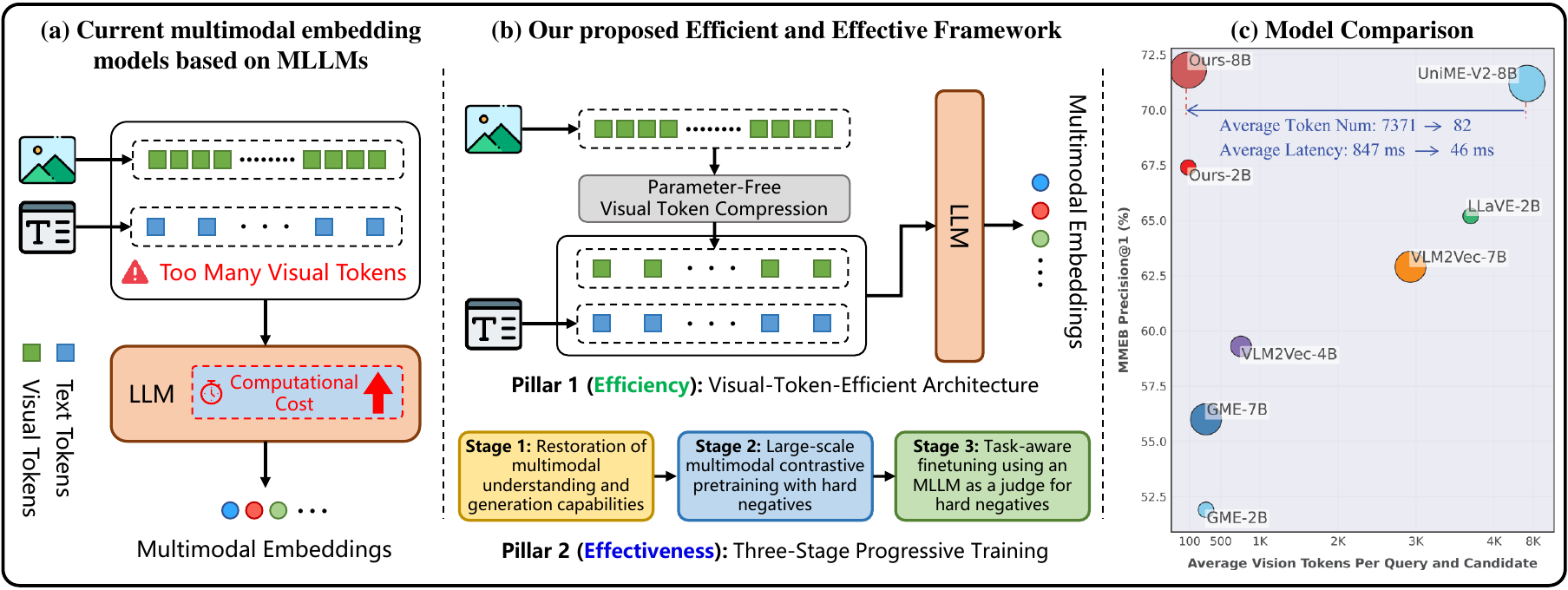}
  \caption{Breaking the efficiency-performance trade-off of MLLM-based embedders for universal multimodal retrieval. (a) MLLM-based embedders suffer from high computational costs due to processing redundant, dense visual token sequences. (b) We propose a visual token compression model paired with a robust three-stage progressive training strategy. (c) Comparisons on MMEB~\cite{VLM2Vec} demonstrate that our approach establishes a new state-of-the-art  using much less visual tokens with reduced inference latency.}
  \label{fig:teaser}
\end{figure}

Building upon these advantages, recent research has rapidly advanced MLLM-based universal multimodal embeddings through enhancement of data scale and quality~\cite{zhang2024gme,zhou2025megapairs,jian2025rzenembed,li2026qwen3}, gradient amplification on hard negatives~\cite{lan2025llave,xue2025improve,jian2025rzenembed}, refinement of hard negative mining strategies~\cite{mm_embed,gu2025break,thirukovalluru2025breaking,jian2025rzenembed,li2026qwen3}, expert knowledge distillation~\cite{gu2025break,gu2026aaai}, multi-stage progressive training~\cite{LamRA,gu2025break,jian2025rzenembed,li2026qwen3}, incorporation of thinking and reinforcement learning~\cite{zhu2025retrv,lan2025ume,cui2025think}, and coordination with reranker during inference~\cite{LamRA,gu2026aaai,li2026qwen3}.

However, these advancements overlook a critical bottleneck: \textit{the prohibitive cost of long visual token sequences}. Standard MLLM architectures typically adopt a dense visual token integration strategy to enhance model performance. For instance, the widely used LLaVA-OneVision~\cite{li2024llavaone} can generate up to 7,290 visual tokens as input to the language model. While this long-sequence injection benefits fine-grained generation tasks like OCR, it introduces massive redundancy for retrieval, where the goal is to distill multimodal information from these redundant visual tokens together with textual tokens into a single \texttt{[EOS]} token. This creates a significant imbalance: the computational cost of processing these redundant visual tokens scales quadratically with their sequence length, while their contribution to  the semantic quality of the final embeddings is often marginal. Consequently, this inefficiency acts as a primary barrier to deploying MLLM-based embedders in large-scale, latency-critical retrieval systems.

To address this challenge of computational inefficiency while delivering high performance, we propose a novel framework that synergistically combines architectural efficiency with a curated training strategy. Architecturally, we adopt a parameter-free spatial interpolation module which projects the long visual sequence into a compressed form, reducing the token overhead by 75\% while avoiding the optimization difficulties of learnable abstractors~\cite{li2023blip,cha2024honeybee}. To mitigate any potential performance degradation from aggressive compression and learn robust discriminative representations, we pair this efficient architecture with a three-stage training pipeline: (1) \textbf{Multimodal Foundational Capability Restoration.} We begin with generative continued training on general multimodal instruction datasets. This stage re-aligns the compressed visual features with the LLM, ensuring the preservation of foundational multimodal understanding and generation abilities. (2) \textbf{Multimodal Contrastive Pretraining.} We construct a general embedder using 16M multimodal samples with contrastive training. This stage aims to cultivate robust general representation capabilities by evolving from a contrastive warm-up to a self-refinement phase with retrieval-based hard negative mining. (3) \textbf{Task-aware Finetuning.} We refine the model on a curated, high-quality multi-task dataset through a retrieve-and-curate strategy. Using the previous stage's model, we retrieve candidates for each query of the training set and leverage an MLLM as a judge to construct high-quality hard negatives. These curated samples then drive the final stage of contrastive learning, yielding  the final embedding model to handle diverse and complex scenarios. Through extensive experiments, our approach establishes a new state-of-the-art on various natural image~\cite{VLM2Vec} and visual document~\cite{meng2025vlm2vec} retrieval tasks. Crucially, this superior performance is achieved with remarkable token efficiency, with only a quarter of the visual tokens, validating the power of our co-designed compression and training strategy. Our main contributions are as follows:

\begin{itemize}
    \item We propose a novel framework that successfully reconciles efficiency and performance for MLLM-based universal embedding. We demonstrate that a model with aggressive visual token compression can significantly outperform its non-compressed counterparts when supported by a dedicated, advanced training pipeline.
    \item  We introduce a coarse-to-fine training strategy designed for compressed MLLMs. This pipeline provides a systematic and effective method for restoring foundational abilities, building robust discriminative power, and achieving strong multi-task generalization with curated data from MLLM-as-a-Judge.
    \item Through extensive experiments, we demonstrate that our proposed Magic-MM-Embedding series establishes new state-of-the-art results, validating the superiority of our holistic approach in developing models that are both computationally efficient and highly effective.
\end{itemize}

\begin{figure}[tb]
  \centering
  \includegraphics[width=\linewidth]{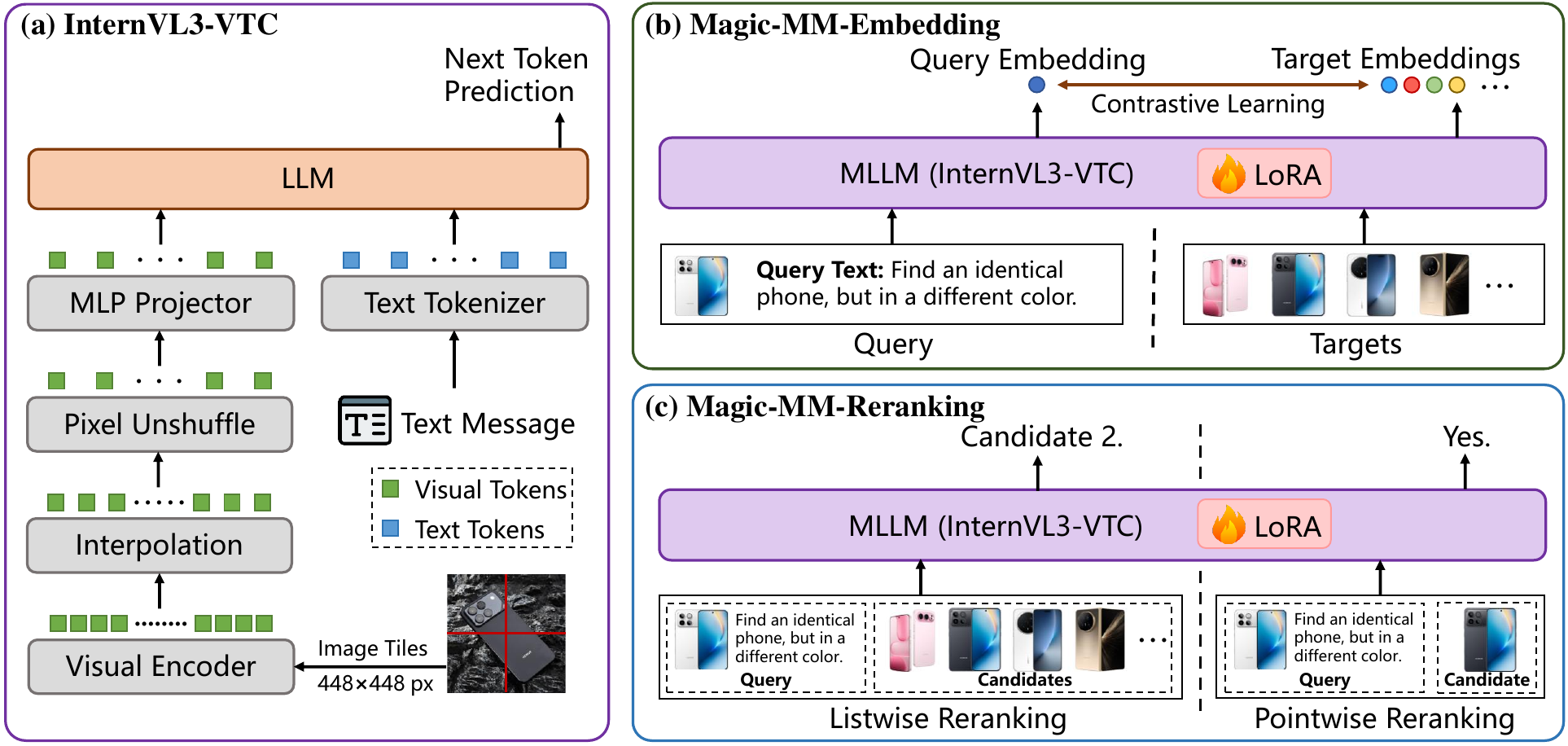}
  \caption{Overview of the proposed visual-token-efficient architecture for universal multimodal retrieval. (a) The proposed MLLM architecture with Visual Token Compression, InternVL3-VTC. (b, c) The proposed inference-efficient, universal multimodal embedder and reranker, both of which are built on InternVL3-VTC.}
  \label{fig:model_arch}
\end{figure}
\section{Related Work}

\noindent{\textbf{CLIP-based Multimodal Embedding. }}Multimodal representation learning was popularized by CLIP-style models~\cite{clip2021Radford,jia2021scaling,li2022blip,zhai2023sigmoid,li2023blip,CLIP_bigG14,EVA_CLIP_18B,longclip}, which adopt an image-text dual-encoder architecture and mainly support bidirectional text-image retrieval. Building on this paradigm, UniIR~\cite{wei2024uniir} and MagicLens~\cite{magiclens} fuse representations from both towers to handle interleaved image-text inputs. However, these methods still follow a late-fusion strategy, encoding each modality independently before fusion, which limits fine-grained cross-modal interaction~\cite{VLM2Vec,E5V,gu2025break}. Moreover, their BERT-style~\cite{devlin2019bert} text encoders suffer from limited real-world knowledge and strict input length constraints, resulting in suboptimal performance on complex text understanding~\cite{longclip,flame,yuksekgonul2022and}.

\noindent{\textbf{MLLM-based Multimodal Embedding. }}Benefiting from the rapid progress of MLLMs~\cite{bai2025qwen3,zhu2025internvl3,li2024llavaone}, which natively support interleaved text-image inputs and strong multimodal understanding, MLLM-based multimodal embedding paradigms have rapidly emerged~\cite{E5V,VLM2Vec}. To improve discriminative capability,  a large number of studies have systematically conducted extensive optimizations, including improving data scale and quality~\cite{zhang2024gme,zhou2025megapairs,jian2025rzenembed,li2026qwen3}, amplifying gradients on hard negatives~\cite{lan2025llave,xue2025improve,jian2025rzenembed}, refining hard negative mining strategies~\cite{mm_embed,gu2025break,thirukovalluru2025breaking,jian2025rzenembed,li2026qwen3}, distilling expert knowledge~\cite{gu2025break,gu2026aaai}, adopting multi-stage progressive training~\cite{LamRA,gu2025break,jian2025rzenembed,li2026qwen3}, introducing thinking and reinforcement learning~\cite{zhu2025retrv,lan2025ume,cui2025think}, and coordinating with reranker during inference~\cite{LamRA,gu2026aaai,li2026qwen3}. With these community efforts, the discriminative power of MLLM-based multimodal embedding models has been significantly improved. However, because these models directly adopt the general-purpose MLLM architecture, the issue of high inference cost~\cite{zhang2025llava,yang2025topv,dhouib2025pact} caused by visual token redundancy has not yet received attention or solutions.

\noindent{\textbf{Visual Token Compression in MLLMs. }}A large body of research has demonstrated that significant visual token redundancy exists in MLLMs~\cite{chen2024image, alvar2025divprune, shang2025llava, yang2025visionthink}. Due to the quadratic computational complexity of the attention mechanism, these redundant tokens substantially reduce inference efficiency. To improve the inference efficiency of MLLMs, prior work has proposed various visual token compression methods~\cite{shao2026surveytokencompressionefficient}. Existing approaches mainly include: transforming feature maps into more compact representations~\cite{chen2024far,zhu2025internvl3,dai2024nvlm,li2024llavaone}, merging redundant tokens based on inter-token similarity~\cite{wang2025folder,alvar2025divprune,yang2025topv}, pruning unimportant tokens according to attention scores~\cite{shang2025llava,yang2025visionzip,zhang2025beyond}, and distilling visual information through learnable queries~\cite{li2023blip,chen2024internvl,bai2023qwenvlversatilevisionlanguagemodel,zhang2025llava}. However, most of these methods are designed for general generation tasks, and their efficiency and effectiveness in multimodal embedding tasks has not yet been systematically studied.

\section{Methodology}

Our objective is to develop a highly efficient and effective MLLM-based universal multimodal embedding model for retrieval.  The overview of the proposed architecture is shown in~\cref{fig:model_arch}. We begin by formally defining the universal multimodal embedding task before detailing our proposed framework, including our architectural modifications, progressive training pipeline and synergistic reranker.

\subsection{Preliminaries}

We formulate the learning of universal multimodal embedding as a unified mapping function within a shared semantic space. Let $\mathcal{X}$ represent the universal multimodal input space. Each input $x \in \mathcal{X}$, which serves as either a query $q$ or a candidate $c$, is composed of task instructions, visual context, and textual context. The corresponding input templates for queries and candidates can be found in Appendix C.1, and the task instructions for different datasets can be found in Appendix D. To obtain multimodal embeddings, we employ an MLLM with visual token compression as the encoder $f : \mathcal{X} \rightarrow \mathbb{R}^{L \times D}$ to map an input $x \in \mathcal{X}$ to a sequence of hidden states $\mathbf{h}_1, \mathbf{h}_2, \ldots, \mathbf{h}_L$, where each hidden state $\mathbf{h}_i \in \mathbb{R}^D$. We apply $\ell_2$ normalization to the hidden representation of the last token, $\mathbf{h}_L$, to obtain the final embedding $\mathbf{z}_x = \frac{\mathbf{h}_L}{\|\mathbf{h}_L\|_2}$.

To learn a discriminative embedding space, we employ the InfoNCE loss~\cite{van2018cpc} for model training. For a given query $q$, we define a candidate set $\mathcal{C}_q=\{c^+_q\} \cup \mathcal{C}^-_q$ for loss calculation, where $c^+_q$ denotes the ground-truth positive target associated with $q$, and $\mathcal{C}^-_q$ is the set of negative targets. Each $c^-_q \in \mathcal{C}^-_q$ represents a negative target obtained via in-batch sampling or hard negative mining.
The model is trained to maximize the semantic alignment between the query and the positive target while suppressing the negatives, by minimizing the following objective:
\begin{equation}\label{eq:infonce}
   \mathcal{L}_{\text{InfoNCE}} = - \log \frac{\exp(\mathbf{z}_q^\top \mathbf{z}_{c^+_q} / \tau)}{ \exp(\mathbf{z}_q^\top \mathbf{z}_{c^+_q} / \tau) + \sum_{c^-_q \in \mathcal{C}^-_q} \exp(\mathbf{z}_q^\top \mathbf{z}_{c^-_q} / \tau)}, 
\end{equation}
where $\tau$ is the temperature, and $(\cdot)^\top (\cdot)$ denotes the dot-product similarity.

\subsection{Parameter-Free Visual Token Compression}  

Given an input image $\mathbf{I} \in \mathcal{I}$, a standard MLLM first applies a visual encoder $e_v : \mathcal{I} \rightarrow \mathbb{R}^{H \times W \times C}$ to produce a spatial feature map $\mathbf{F} \in \mathbb{R}^{H \times W \times C}$, which is conventionally flattened into $H \times W$ visual tokens and projected into the LLM. However, the resulting long token sequence incurs a substantial computational overhead due to the quadratic complexity of self-attention. To mitigate this issue, we insert a \textit{parameter-free visual compression module} after the visual encoder that applies bilinear interpolation to downsample the spatial dimensions of $\mathbf{F}$ via an operator $\Phi$, yielding a compressed feature map $\mathbf{F}' = \Phi(\mathbf{F}) \in \mathbb{R}^{H' \times W' \times C}$. The compressed map $\mathbf{F}'$ is further downsampled via the pixel unshuffle operation $\Psi$ with a downscaling factor $\alpha$: $\mathbf{F''}=\Psi(\mathbf{F'}, \alpha)\in \mathbb{R}^{\frac{H'}{\alpha} \times \frac{W'}{\alpha} \times \alpha^2 C}$~\cite{zhu2025internvl3}. $\mathbf{F''}$ is then flattened and fed into the LLM, reducing the number of visual tokens and inference cost without introducing parameters. Details are in Appendix A.

\subsection{Progressive Coarse-to-Fine Training Pipeline}

Our visual token compression architecture addresses the high attention computation cost in MLLM-based embedders. We then introduce a progressive coarse-to-fine training pipeline to recover and further enhance performance, enabling the model to achieve both efficiency and effectiveness.

\noindent\textbf{Stage 1: Multimodal Foundational Capability Restoration.} The introduction of the interpolation module alters the spatial structure and density of visual features that the pretrained LLM backbone expects.
Therefore, the primary goal of the first stage is not retrieval, but alignment. 
To this end, we re-align compressed visual representations with the LLM semantic space. Through generative training on general-purpose multimodal instruction-following datasets, we restore fundamental multimodal understanding and generation capabilities.
For the textual response token sequence $y_1, y_2, \dots, y_T$, the model is optimized using the standard auto-regressive Next Token Prediction (NTP) loss:
\begin{equation}
    \mathcal{L}_{\text{NTP}} = - \sum_{t=1}^{T} \log P(y_t | y_{<t}, x), 
\end{equation}
where $y_t$ is the ground-truth next token, and $x \in \mathcal{X}$ denotes the multimodal input consisting of visual and textual context. 
This step is vital to bridge the distribution gap caused by token compression, ensuring the MLLM retains its reasoning capabilities before transitioning to embedding learning.

\noindent\textbf{Stage 2: Multimodal Contrastive Pretraining.} With the multimodal foundational ability restored, we pivot the model towards multimodal representation learning. This stage operates on a large-scale multimodal retrieval corpus and proceeds in two steps to progressively increase difficulty. We first warm up the model by training with standard InfoNCE loss using in-batch negatives as Eq.~\eqref{eq:infonce}. Subsequently, to encourage the model to learn fine-grained distinctions, we introduce a \textit{Global Hard Negative Mining} strategy to inject hard negatives into the training set and conduct a new round of training. Unlike the warm-up phase, which uses random in-batch negatives, for each sub-dataset, we mine informative negatives for every query from all candidates in the entire dataset. Specifically, for each query $q$, we retrieve a ranked list of candidates. We exclude the ground-truth positive $c^+_q$ from this list and randomly sample 2 hard negatives. These negatives are selected from positions 50–100 in the list, which helps avoid false negatives that are common in top-ranked candidates (e.g., top-10), while keeping the negatives more challenging than random batch negatives.

\noindent\textbf{Stage 3: Task-Aware Finetuning with an MLLM as a Judge.} The final stage focuses on enhancing the model for handling diverse scenarios and complex tasks. Standard training datasets often suffer from ``false negatives'' and lack sufficiently challenging negatives. To resolve this, we further employ an expert MLLM as a judge to perform data curation and generate high-fidelity hard negatives. Concretely, for each query $q$ in the target training set, we perform a ``retrieve-and-judge'' process. We first utilize our model from stage 2 to retrieve the top-$K$ ($K=20$) candidates $\mathcal{C}_{\text{ret}}$. We feed each pair $(q, c_i)$, where $c_i \in \mathcal{C}_{\text{ret}}$, into Qwen3-VL~\cite{bai2025qwen3} to assess their relevance. The judgment template can be found in  Appendix C.1, and the judgment instructions used for each dataset are presented in Appendix E. We then examine the output logits of the \texttt{`yes'} and \texttt{`no'} tokens to determine relevance. If $\text{logit}(\text{\texttt{yes}}) > \text{logit}(\text{\texttt{no}})$, the candidate is deemed relevant. This helps us discover previously unlabeled true positives, thereby expanding the positive set beyond the original ground-truth positives. If $\text{logit}(\text{\texttt{no}}) > \text{logit}(\text{\texttt{yes}})$, the candidate is deemed irrelevant. Since these items were retrieved in the top-20 by our model, they constitute high-quality hard negatives. For contrastive loss calculation, we keep the original ground-truth $c^+_q$ as the only positive to preserve consistency. The negative sample set $\mathcal{C}^-_q$ is augmented with judge-identified hard negatives, which serve as challenging distractors and force the model to distinguish more confusing samples.

\subsection{Synergistic Reranker}

To construct a comprehensive retrieval system following previous works~\cite{LamRA,gu2026aaai}, we train a reranker based on the model from stage 1 to leverage its preserved multimodal understanding and generation capabilities. Thanks to the flexibility of MLLM models, we employ a joint training strategy that combines pointwise and listwise training objectives. Crucially, unlike standard approaches that rely solely on the original dataset labels, our reranker is trained on the judge-curated set from stage 3 of embedding model training. For a given query $q$, we define the augmented positive set $\mathcal{C}_{\text{aug}}^+=\{c_q^+\} \cup \mathcal{C}_{\text{judge}}^+$ as the union of the original ground-truth $c_q^+$ and the judge-identified true positives $\mathcal{C}_{\text{judge}}^+$. Similarly, the negative set $\mathcal{C}_{\text{judge}}^-$ consists of the judge-verified hard negatives.

For the pointwise reranking formulation, the model evaluates query-candidate pairs independently. We construct training triplets by sampling a positive candidate $c^+ \in \mathcal{C}_{\text{aug}}^+$ and a hard negative candidate $c^- \in \mathcal{C}_{\text{judge}}^-$. We instruct the model to output the token \texttt{`Yes'} for positive pairs and \texttt{`No'} for negative pairs. The instruction template for pointwise reranking can be found in Appendix C.2. The pointwise loss is minimized using standard Cross Entropy (CE) loss:
\begin{equation}
    \mathcal{L}_{\text{point}} = \mathcal{L}_{\text{CE}}(\texttt{Yes}, r(q, c^+)) + \mathcal{L}_{\text{CE}}(\texttt{No}, r(q, c^-)),
\end{equation}
where $r(\cdot)$ represents the autoregressive reranker. 

For listwise reranking training, we construct a candidate list by sampling $M$ hard negatives $c^-_1, \dots, c^-_M$ from $\mathcal{C}_{\text{judge}}^-$ (where $M \in \{2,3,4,5\}$) and one positive candidate $c^+$ sampled from the augmented set $\mathcal{C}_{\text{aug}}^+$.
We randomly insert $c^+$ into the list at position $k$ and prompt the model to identify the most relevant candidate. The input template for listwise reranking can be found in Appendix C.2. The model is trained to directly generate the position index $k$ of the positive candidate. The listwise loss is formulated as:
\begin{equation}
\mathcal{L}_{\text{list}} = \mathcal{L}_{\text{CE}}(k, r(q, c^-_1, \dots, c^+, \dots, c^-_M))
\end{equation}
The final objective combines both tasks: $\mathcal{L}_{\text{total}} = \mathcal{L}_{\text{point}} + \mathcal{L}_{\text{list}}$.
\section{Dataset Construction}

\noindent{\textbf{Stage 1. }}In this stage, to restore the multimodal understanding and generation abilities of the token compression model, we construct a multimodal instruction-following dataset containing 32M examples. This corpus is composed of both open-source (26.2M) and in-house data (5.8M), covering various tasks, including multimodal and text-only instruction data, captioning, grounding and classification. A detailed breakdown of the dataset is in Appendix F. We apply rule-based deduplication and standardize all annotations into a unified format.

\noindent{\textbf{Stage 2. }}In this stage, to adapt the model to multimodal representation learning and strengthen its discriminative power, we construct a multimodal retrieval dataset comprising 16M samples collected from open-source datasets. The training data includes three categories: single-modal, cross-modal, and fused-modal, and is sampled from MegaPairs~\cite{zhou2025megapairs}, Colpali train set~\cite{faysse2024colpali}, VisRAG~\cite{VisRAG}, Docmatix~\cite{laurenccon2024building}, BAAI-MTP~\cite{baai-mtp}, ImageNet-1K~\cite{deng2009imagenet}, BLIP Bootstrapped Image-Text Pairs~\cite{li2022blip}, MMEB-train~\cite{VLM2Vec}, and mmE5-synthetic~\cite{chen2025mme5}. A more detailed description of the stage 2 training data is provided in Appendix F.

\noindent{\textbf{Stage 3. }}At this stage, to enhance the model’s ability to handle diverse and complex scenarios, we curated a dataset containing 1.5M high-quality, multi-task samples. These data are designed for both natural image retrieval tasks and visual document retrieval tasks. For the natural image retrieval tasks, we use MMEB-train~\cite{VLM2Vec} as the training set, while for the visual document retrieval tasks, we adopt the Colpali train set~\cite{faysse2024colpali} and VisRAG~\cite{VisRAG} as training data.
\section{Experiments}

\subsection{Evaluation Setup \& Benchmarks}

We first evaluate the performance of Magic-MM-Embedding on natural image retrieval and visual document retrieval tasks. For natural image retrieval, we use MMEB~\cite{VLM2Vec}, a comprehensive benchmark comprising 36 sub-datasets and 4 meta-tasks, to assess and report Precision@1. For visual document retrieval (VisDoc), we follow the VLM2Vec-V2~\cite{meng2025vlm2vec} settings and use ViDoRe v1 (VDRv1)~\cite{faysse2024colpali}, ViDoRe v2 (VDRv2)~\cite{mace2025vidore}, VisRAG (VR)~\cite{VisRAG}, and ViDoSeek~\cite{wang2025vidorag}+MMLongBench-Doc (OOD)~\cite{ma2024mmlongbench} to evaluate and report NDCG@5. To assess the performance of Magic-MM-Embedding on cross-modal retrieval, following the UniME-V2 settings~\cite{gu2026aaai}, we further conduct evaluations on Flickr30K~\cite{plummer2015flickr30k}, MSCOCO~\cite{lin2014microsoft}, ShareGPT4V~\cite{chen2024sharegpt4v}, Urban1K~\cite{longclip}, and SugarCrepe~\cite{hsieh2023sugarcrepe} and report Precision@1. 

\subsection{Implementation Details}

\noindent\textbf{Model Architecture.} We implement our framework using PyTorch and the ms-swift~\cite{zhao2024swiftascalablelightweightinfrastructure} library. We adopt InternVL3~\cite{zhu2025internvl3} as our backbone MLLM, referring to our proposed Visual Token Compression variant as~\textbf{InternVL3-VTC}. Specifically, for the feature map produced from each image tile, we utilize bilinear interpolation (Discussion of different compression methods is in Appendix B.1.) to downsample the number of visual tokens to 16 $\times$ 16 = 256. After applying pixel unshuffle, we finally obtain 8 $\times$ 8 = 64 tokens as input to the LLM, thereby retaining only one-fourth of the original visual tokens. See Appendix B.2 for the ablation study of multimodal understanding ability vs. the number of visual tokens.

\noindent\textbf{Image Tiling Strategy.} For the image tiling strategy, we follow the approach used in InternVL3~\cite{zhu2025internvl3}. To ensure computational efficiency, however, we reduce the maximum number of image tiles (\texttt{MAX\_NUM}) during both training and inference. Specifically, in stage 1 training, \texttt{MAX\_NUM} is set to 4. During the training and inference of downstream embedder and reranker models, we adopt a data-dependent policy: for data containing visual document images, \texttt{MAX\_NUM} is set to 4, while for all other natural image data, \texttt{MAX\_NUM} is uniformly set to 1.

\noindent\textbf{Implementation of the Embedder and Reranker.} The embedder and reranker are trained in a multi-stage manner. In stage 1, we train all model parameters to restore the multimodal understanding and generative capabilities of the model. Stages 2 and 3 adopt contrastive pretraining and task-aware fine-tuning with LoRA, using a unified LoRA rank of 16. Discussion of the LoRA rank of the embedder can be found in Appendix B.3. We use Qwen3-VL-8B~\cite{bai2025qwen3} as the judge in the stage 3 and insert 12 hard negatives per training instance. The reranker is initialized from the stage 1 checkpoint and trained on the same judge-curated data as stage 3, using a combination of pointwise and listwise objectives. During inference, we perform two-stage retrieval: first, the embedder retrieves a candidate set, followed by pointwise reranking of the top-5 candidates. Detailed training configurations are provided in Appendix G.

\begin{table}[t]
\centering
\caption{Results on the MMEB benchmark~\cite{VLM2Vec}. The scores are averaged per meta-task. The best performance in each block is in \textbf{bold}. ``E'' refers to the single-stage retrieval performance using only embedder; ``E+R'' refers to the two-stage retrieval results, obtained by first using the embedder to retrieve a candidate set, followed by a final ranking with the reranker.
}
\resizebox{\linewidth}{!}{
\begin{tabular}{lrccccccccc}
\toprule
\multirow{2}{*}{\textbf{Model}} & \multirow{2}{*}{\textbf{Backbone (Model Size)}}  & \multicolumn{4}{c}{\textbf{Per Meta-Task Score}} & & \multicolumn{3}{c}{\textbf{Average Score}} \\ 
\cmidrule(lr){3-6} \cmidrule(lr){8-10}
                       & & \textbf{Classification} & \textbf{VQA}  & \textbf{Retrieval} & \textbf{Grounding} & & \textbf{IND} & \textbf{OOD} & \textbf{Overall} \\ \midrule
\textbf{\# of datasets} $\rightarrow$ & & 10 & 10 & 12 & 4 & & 20 & 16 & 36 \\ \midrule
\multicolumn{10}{c}{\it Zero-shot Results} \\
\midrule
CLIP~\cite{clip2021Radford} &  - (0.4B) & 42.8 & 9.1 & 53.0 & 51.8 && 37.1 & 38.7 & 37.8 \\ 
SigLIP~\cite{zhai2023sigmoid} & - (0.9B) & 40.3 & 8.4 & 31.6 & 59.5 && 32.3 & 38.0 & 34.8\\
EVA-CLIP~\cite{EVA_CLIP_18B} & - (8.1B) & 56.0 & 10.4 & 49.2 & 58.9 & & 38.1 & 45.6 & \textbf{43.7} \\
MagicLens~\cite{magiclens} & - (0.4B) & 38.8 & 8.3 & 35.4 & 26.0 && 31.0 & 23.7 & 27.8 \\
E5-V~\cite{E5V} & Phi3.5-V (4.2B) & 39.1 & 9.6 & 38.0 & 57.6 && 33.1 & 31.9 & 36.1 \\
E5-V~\cite{E5V} & LLaVA-1.6 (8.4B) & 39.7 & 10.8 & 39.4 & 60.2 && 34.2 & 33.4 & 37.5 \\

\midrule
\multicolumn{10}{c}{\textit{Trained with MMEB}} \\
\midrule
VLM2Vec-V1~\cite{VLM2Vec} & Qwen2-VL (2.2B) & 59.0 & 49.4 & 65.4 & 73.4 && 66.0 & 52.6 & 59.3 \\
UniME~\cite{gu2025break} & Phi3.5-V (4.2B) & 54.8 & 55.9 & 64.5 & 81.8 && 68.2 & 52.7 & 64.2 \\
LLaVE~\cite{lan2025llave} & Aquila-VL (2.0B) & 62.1 & 60.2 & 65.2 & 84.9 && 69.4 & 59.8 & 65.2 \\
UniME-V2 (E)~\cite{gu2026aaai} & Qwen2-VL (2.2B) & 62.1 & 56.3  & 68.0 & 72.7 & & 67.4 & 58.9 & 63.6 \\
UniME-V2 (E+R)~\cite{gu2026aaai} & Qwen2-VL (2.2B) & 64.1 & 64.3 & 71.6 & 70.6 && 69.8 & 64.3 & 67.4\\
\rowcolor{brown!10} \textbf{Magic-MM-Embedding (E)}   & InternVL3-VTC (1.9B) & 60.9 & 63.3 & 72.2 & 84.6 && 74.7 & 59.5 & 68.0  \\
\rowcolor{brown!20} \textbf{Magic-MM-Embedding (E+R)}   & InternVL3-VTC (1.9B) & 61.3 & 67.2 & 73.5 & 89.8 &&75.2 & 63.9 & \textbf{70.2} \\
\hdashline
VLM2Vec-V1~\cite{VLM2Vec} & Qwen2-VL (8.3B) & 62.6 & 57.8 & 69.9 & 81.7 && 65.2 & 56.3 & 65.8 \\
UniME~\cite{gu2025break} & LLaVA-OV (8.0B) & 66.8 & 66.6 & 70.5 & 90.9 && 74.6 & 65.8 & 70.7 \\
LLaVE~\cite{lan2025llave} & LLaVA-OV (8.0B) & 65.7 & 65.4 & 70.9 & 91.9 && 75.0 & 64.4 & 70.3 \\
QQMM~\cite{xue2025improve}   & LLaVA-OV  (8.0B)    & 66.8 & 66.8 & 70.5 & 90.4 && 74.7 & 65.6 & 70.7 \\
UniME-V2 (E)\cite{gu2026aaai} & LLaVA-OV (8.0B) & 65.3  & 67.6  & 72.9  & 90.2 && 74.8 & 66.7 & 71.2 \\
UniME-V2 (E) \cite{gu2026aaai} & Qwen2-VL (8.3B) & 64.0 &  60.1&  73.1 & 82.8 && 72.0 & 63.0 & 68.0  \\
UniME-V2 (E+R) \cite{gu2026aaai} & Qwen2-VL (8.3B) & 63.8 & 66.3 & 73.5 & 75.0 && 71.7 & 65.6 & 69.0  \\
\rowcolor{brown!10} \textbf{Magic-MM-Embedding (E)} & InternVL3-VTC (8.1B) & 65.0 & 68.2 & 74.7 & 89.6 && 78.4 & 63.7 & 71.9 \\
\rowcolor{brown!20} \textbf{Magic-MM-Embedding (E+R)} & InternVL3-VTC (8.1B) & 64.4 & 71.0 & 75.7 & 90.1 &&78.4 & 65.9 & \textbf{72.9} \\

\bottomrule
\end{tabular}}

\label{tab:mmeb}
\end{table}

\subsection{Experimental Results}

\noindent\textbf{Multimodal Retrieval on MMEB.} In~\cref{tab:mmeb}, we present the performance comparison against representative baselines. As the results show, dual-tower models such as CLIP significantly lag behind MLLM-based approaches due to the inherent limitations of the architecture. Among the MLLM methods, our proposed embedder establishes a new state-of-the-art, surpassing strong baselines like UniME-V2~\cite{gu2026aaai} and QQMM~\cite{xue2025improve}, proving that our progressive training strategy overcomes the potential information loss of token compression. For the ``Embedder + Reranker (E+R)'' paradigm, both our method and UniME-V2~\cite{gu2026aaai} show that adding a reranker further boosts accuracy. However, our approach consistently performs better: at the 2B and 8B scales, our models outperform UniME‑V2 in both the embedder-only setting and the full ``E+R'' setting. Notably, our embedder‑only setting outperforms UniME‑V2’s ``E+R'' configuration. This confirms that our framework provides a stronger foundational retriever and a more effective overall pipeline than the previous best-performing method.

\begin{table}[!htpb]
\centering
\setlength\tabcolsep{7pt}
\caption{Results on VisDoc~\cite{meng2025vlm2vec}. The best performance in each block is in \textbf{bold}. ``E'' refers to the single-stage retrieval performance using only embedder; ``E+R'' refers to the two-stage retrieval results, obtained by first using the embedder to retrieve a candidate set, followed by a final ranking from the reranker.}
\renewcommand{\arraystretch}{1.2}
\resizebox{0.85\textwidth}{!}{
\begin{tabular}{lrccccc }
\toprule
\multirow{2}{*}{\textbf{Model}} & \multirow{2}{*}{\textbf{Backbone (Model Size)}}  & \multicolumn{5}{c}{\textbf{VisDoc}}\\
\cmidrule(lr){3-7} 
&  & \textbf{VDRv1} & \textbf{VDRv2} & \textbf{VR} & \textbf{OOD} & \textbf{Overall} \\
\midrule
\textbf{\# of Datasets} $\rightarrow$ & & 10 & 4 & 6 & 4 & 24\\
\midrule
GME~\cite{zhang2024gme} & Qwen2-VL (2.2B) & 86.1 & 54.0 & 82.5 & 43.1 & 72.7\\
ColPali~\cite{faysse2024colpali} & Paligemma (2.9B) & 83.6 & 52.0 & 81.1 & 43.1 & 71.0\\
Ops-MM-embedding-v1~\cite{OpsMMembeddingv1}  & Qwen2-VL (8.3B) & 80.1 & 59.6  &79.3 & 43.3 & 70.3\\
VLM2Vec-V2~\cite{meng2025vlm2vec} & Qwen2-VL (2.2B) &75.5 & 44.9 & 79.4 & 39.4 & 65.4\\
\rowcolor{brown!10} \textbf{Magic-MM-Embedding (E)}   & InternVL3-VTC (1.9B) & 83.4 & 53.3 & 85.6 & 42.2 & 72.1\\
\rowcolor{brown!20} \textbf{Magic-MM-Embedding (E+R)}   & InternVL3-VTC (1.9B) & 84.4 & 56.1 & 87.4 & 41.8 & \textbf{73.3}\\
\hdashline
Ops-MM-embedding-v1~\cite{OpsMMembeddingv1} & Qwen2-VL (8.3B) & 80.1 & 59.6  &79.3 & 43.3 & 70.3\\
GME~\cite{zhang2024gme} & Qwen2-VL (8.3B) & 89.4 & 55.6 & 85.0 & 44.4 & 75.2\\
LamRA-Qwen2~\cite{LamRA} & Qwen2-VL (8.3B) & 22.0 & 11.5 & 37.4 & 21.0 & 23.9\\
LamRA-Qwen2.5~\cite{LamRA} & Qwen2.5-VL (8.3B) & 56.3 & 33.3 & 58.2 & 40.1 & 50.2\\
VLM2Vec-V2~\cite{meng2025vlm2vec} & Qwen2-VL (8.3B) & 78.8 & 52.6 & 82.7 & 42.1 & 69.3\\
\rowcolor{brown!10} \textbf{Magic-MM-Embedding (E)} & InternVL3-VTC (8.1B) & 85.9 & 59.9 & 87.4 & 43.5 & 74.9\\
\rowcolor{brown!20} \textbf{Magic-MM-Embedding (E+R)} & InternVL3-VTC (8.1B) & 86.8 & 59.6 & 89.1 & 42.9 & \textbf{75.5}\\
\bottomrule
\end{tabular}}
\label{tab:visdoc}
\end{table}

\noindent\textbf{Multimodal Retrieval on VisDoc.} Beyond general multimodal retrieval, we evaluate our model on the challenging domain of Visual Document Retrieval (VisDoc), a fine-grained task demanding high-resolution inputs to preserve textual details. In~\cref{tab:visdoc}, we report results. Surprisingly, despite compressing visual tokens by 75\%, our token-efficient method achieves state-of-the-art results, challenging the assumption that high redundancy is strictly necessary for fine-grained retrieval. Additionally, we observe that GME~\cite{zhang2024gme} is a formidable baseline, outperforming our standalone embedder; we attribute this to GME's use of massive, proprietary task-aware datasets compared to our smaller, publicly available training sources. However, our full ``Embedder + Reranker'' pipeline surpasses GME to establish a new state-of-the-art. This shows that our synergistic training strategy effectively bridges the data gap, allowing us to achieve superior performance using only public data and significantly fewer visual tokens.

\begin{table}[H]
\centering
\caption{Cross-modal retrieval results on Flickr30K~\cite{plummer2015flickr30k}, MSCOCO~\cite{lin2014microsoft}, ShareGPT4V~\cite{chen2024sharegpt4v}, Urban1K~\cite{longclip} and SugarCrepe~\cite{hsieh2023sugarcrepe}.}
\setlength\tabcolsep{0.75pt}
\renewcommand\arraystretch{1.3}
\fontsize{7pt}{7pt}\selectfont
\resizebox{0.95\linewidth}{!}{
\begin{tabular}{@{}l r cc cc cc cc ccc@{}}
\toprule
\multicolumn{1}{l}{\multirow{4}{*}{\textbf{Models}}}
& \multicolumn{1}{c}{\multirow{4}{*}{\textbf{Backbone (Model Size)}}}
& \multicolumn{4}{c}{\textbf{Short Caption}} 
& \multicolumn{4}{c}{\textbf{Long Caption}} 
& \multicolumn{3}{c}{\textbf{Compositional}} \\
\cmidrule(lr){3-6} \cmidrule(lr){7-10} \cmidrule(l){11-13} 

& & \multicolumn{2}{c}{\textbf{Flickr30K}} & \multicolumn{2}{c}{\textbf{MSCOCO}} 
& \multicolumn{2}{c}{\textbf{ShareGPT4V}} & \multicolumn{2}{c}{\textbf{Urban1K}} 
& \multicolumn{3}{c}{\textbf{SugarCrepe}} \\
\cmidrule(lr){3-4} \cmidrule(lr){5-6} \cmidrule(lr){7-8} \cmidrule(lr){9-10} \cmidrule(l){11-13}

& & $\text{T}\rightarrow \text{I}$ & $\text{I}\rightarrow \text{T}$ & $\text{T}\rightarrow \text{I}$ & $\text{I}\rightarrow \text{T}$ 
& $\text{T}\rightarrow \text{I}$ & $\text{I}\rightarrow \text{T}$ & $\text{T}\rightarrow \text{I}$ & $\text{I}\rightarrow \text{T}$ 
& Replace & Swap & Add \\
\midrule

OpenCLIP~\cite{clip2021Radford} & - (0.4B) & 67.3 & 87.2 & 37.0 & 58.1 & 81.8 & 84.0 & 47.0 & 47.0 & 79.5 & 62.7 & 74.9 \\ 
CLIP~\cite{CLIP_bigG14} & - (2.5B) & 79.5 & 92.9 & 51.3 & 67.3 & 90.1 & 93.6 & 77.8 & 80.7 & 86.5 & 68.9 & 88.4 \\
EVA-CLIP~\cite{EVA_CLIP_18B} & - (8.1B) & 80.3 & \textbf{94.5} & 52.0 & 70.1 & 93.1 & 91.2 & 80.4 & 77.8 & 85.9 & 70.3 & 86.7 \\
\hdashline
E5-V~\cite{E5V} & Phi3.5-V (4.2B) & 72.2 & 79.6 & 44.7 & 53.4 & 86.0 & 88.5 & 83.8 & 83.6 & 88.2 & 66.6 & 75.3 \\
VLM2Vec~\cite{VLM2Vec} & Qwen2-VL (2.2B) & 69.3 & 89.6 & 40.0 & 62.5 & 78.1 & 88.2 & 78.7 & 83.9 & 67.2 & 46.5 & 66.4 \\
UniME~\cite{gu2025break} & Qwen2-VL (2.2B) & 74.9 & 90.6 & 44.0 & 63.5 & 83.6 & 88.6 & 83.3 & 83.2 & 65.6 & 45.2 & 65.7 \\
UniME-V2~\cite{gu2026aaai} & Qwen2-VL (2.2B) & 79.8 & 89.9 & 53.7 & 65.1 & 91.6 & 94.2 & 95.6 & 92.2 & 70.9 & 51.2 & 70.2 \\
\rowcolor{brown!20}
\textbf{Magic-MM-Embedding} & InternVL3-VTC (1.9B) & \textbf{84.4} & 93.0 & \textbf{61.4} & \textbf{75.8} & \textbf{97.2} & \textbf{97.3} & \textbf{98.4} & \textbf{97.8} & \textbf{91.6} & \textbf{82.6} & \textbf{94.2} \\
\hdashline
E5-V~\cite{E5V} & LLaVA-1.6 (8.4B) & 77.3 & 85.7 & 49.1 & 57.6 & 85.1 & 82.1 & 88.9 & 83.2 & 86.3 & 68.7 & 66.9 \\
VLM2Vec~\cite{VLM2Vec} & Qwen2-VL (8.3B) & 80.0 & 94.2 & 49.2 & 68.5 & 78.5 & 90.4 & 94.0 & 94.2 & 70.0 & 51.7 & 72.2 \\
UniME~\cite{gu2025break} & Qwen2-VL (8.3B) & 80.8 & 92.7 & 50.9 & 69.8 & 86.5 & 93.8 & 95.3 & 94.0 & 68.8 & 53.0 & 69.8 \\
UniME~\cite{gu2025break} & LLaVA-OV (8.0B) & 83.3 & 94.4 & 54.8 & 74.0 & 93.9 & 89.3 & 94.3 & 95.5 & 80.5 & 65.5 & 82.2 \\
UniME-V2~\cite{gu2026aaai}  & Qwen2-VL (8.3B) & 84.6 & 93.5 & 57.3 & 70.3 & 94.3 & 95.2 & 97.2 & 96.3 & 77.8 & 62.2 & 79.0 \\
UniME-V2~\cite{gu2026aaai}  & LLaVA-OV (8.0B) & \textbf{85.5} & 93.7 & 60.9 & 74.1 & 95.1 & 94.1 & 96.3 & 96.7 & 88.6 & 73.7 & 90.5 \\

\rowcolor{brown!20}
\textbf{Magic-MM-Embedding} & InternVL3-VTC (8.1B) & 82.9 & 94.3 & \textbf{62.9} & \textbf{78.4} & \textbf{98.5} & \textbf{98.3} & \textbf{98.3} & \textbf{98.4} & \textbf{93.2} & \textbf{88.2} & \textbf{95.6} \\
\bottomrule
\end{tabular}
}

\label{tab:cross_modality_retrieval}
\end{table}

\noindent\textbf{Text-Image Cross-Modal Retrieval.} Following previous works~\cite{gu2026aaai, gu2025break}, we further evaluate the text-image cross-modal retrieval ability of our embedding model without reranker. Based on the results in~\cref{tab:cross_modality_retrieval}, in almost all datasets, our embedding method consistently delivers new state-of-the-art results. On the 8B scale, our method achieves significant gains even on benchmarks where the strong baseline UniME-V2 has reached near-saturation levels of 95.0. Specifically, we improve Precision@1 on ShareGPT4V (text-to-image) from 95.1 to 98.5 and on Urban1K (image-to-text) from 96.7 to 98.4. The superiority is even more profound at the 2B scale, where our model dominates UniME-V2 on the challenging SugarCrepe benchmark by a massive margin—scoring 91.6, 82.6, and 94.2 on its three sub-settings compared to 70.9, 51.2, and 70.2, respectively. Crucially, these results of our method are achieved using only 64 visual tokens, far fewer than other standard methods. This empirical evidence leads to a pivotal conclusion: visual token compression is not a trade-off but a strategic advantage. When synergized with our progressive training pipeline, it significantly enhances inference efficiency while simultaneously achieving better cross-modal alignment.

\begin{table}[H]
    \centering
    \setlength\tabcolsep{4pt}
    \caption{Inference efficiency comparison. \textbf{\#$AVT_q$} and \textbf{\#$AVT_c$} refer to the Average number of Visual Tokens in queries and candidates containing images, respectively. \textbf{$l_q$} and \textbf{$l_c$} mean the average latency (millisecond) of query inference and candidate inference, respectively. The best performance in each block is in \textbf{bold}.}
    \label{tab:infer_efficiency}
    \resizebox{\linewidth}{!}{
        \begin{tabular}{lrc>{\centering\arraybackslash}p{1.4cm}c>{\centering\arraybackslash}p{1.4cm}c>{\centering\arraybackslash}p{1.4cm}c>{\centering\arraybackslash}p{1.4cm}}
            \toprule
            \multirow{2}{*}{\textbf{Model}} & \multirow{2}{*}{\textbf{Backbone (Model Size)}} & \multicolumn{4}{c}{\textbf{MMEB}} & \multicolumn{4}{c}{\textbf{VisDoc}} \\
            \cmidrule(lr){3-6} \cmidrule(lr){7-10}
            & & \textbf{\#$AVT_q$} & \textbf{$l_q$} & \textbf{\#$AVT_c$} & \textbf{$l_c$} & \textbf{\#$AVT_q$} & \textbf{$l_q$} & \textbf{\#$AVT_c$} & \textbf{$l_c$} \\
            \midrule
            VLM2Vec~\cite{VLM2Vec} & Phi3.5-V (4.2B) & 757.0 & 99.4 & 757.0 & 85.9 & 0 & 34.0 & 757.0 & 128.6 \\
            GME~\cite{zhang2024gme} & Qwen2-VL (2.2B) & 362.8 & 46.8 & 256.0 & 34.5 & 0 & 19.3 & 1024.0 & 153.8 \\
            LLaVE~\cite{lan2025llave} & Aquila-VL (2.0B) & 3699.0 & 162.8 & 3699.0 & 143.0 & 0 & \textbf{18.5} & 3699.0 & 233.6 \\
            InternVL3~\cite{zhu2025internvl3} & InternVL3 (1.9B) & 398.4 & 37.1 & 256.0 & 29.2 & 0 & 19.8 & 1280.0 & 103.6 \\
            \rowcolor{brown!20} \textbf{Magic-MM-Embedding} & InternVL3-VTC (1.9B) & \textbf{99.6} & \textbf{29.9} & \textbf{64.0} & \textbf{26.1} & 0 & 19.7 & \textbf{320.0} & \textbf{57.3} \\
            \hdashline
            VLM2Vec~\cite{VLM2Vec} & LLaVA-1.6 (8.4B) & 2928.0 & 332.3 & 2928.0 & 278.9 & 0 & 32.4 & 2928.0 & 458.1 \\
            GME~\cite{zhang2024gme} & Qwen2-VL (8.3B) & 362.8 & 82.2 & 256.0 & 56.7 & 0 & \textbf{26.6} & 1024.0 & 268.2 \\
            LamRA~\cite{LamRA} & Qwen2.5-VL (8.3B) & 362.8 & 83.4 & 256.0 & 61.6 & 0 & 28.9 & 1024.0 & 251.7 \\
            UniME-V2~\cite{gu2026aaai} & LLaVA-OV (8.0B) & 7371.0 & 906.9 & 7371.0 & 788.1 & 0 & 32.1 & 7371.0 & 1341.1 \\
            InternVL3~\cite{zhu2025internvl3} & InternVL3 (8.1B) & 398.4 & 76.7 & 256.0 & 55.9 & 0 & 33.8 & 1280.0 & 260.4 \\
            \rowcolor{brown!20} \textbf{Magic-MM-Embedding} & InternVL3-VTC (8.1B) & \textbf{99.6} & \textbf{50.9} & \textbf{64.0} & \textbf{40.6} & 0 & 33.8 & \textbf{320.0} & \textbf{94.8} \\
            \bottomrule
        \end{tabular}
    }
\end{table}

\noindent\textbf{Comparison on Inference Cost.} We compared the inference efficiency of the proposed Magic-MM-Embedding with the representative MLLM-based embedders, as shown in~\cref{tab:infer_efficiency}. For each MLLM backbone, one embedding model was selected. We randomly sampled 5,000 queries and candidates from the MMEB and VisDoc training sets. We measured the average inference latency and the average number of visual tokens for queries and candidates on both MMEB and VisDoc. To ensure fairness, natural images were resized to 448$\times$448, and visual document images were resized to 896$\times$896. All results were obtained using an NVIDIA L20 (48GB) GPU with a batch size of 1 and BF16 precision. No acceleration techniques were used during testing.

Compared with models of similar parameter scales, Magic-MM-Embedding demonstrated significantly lower inference latency than almost all existing models, thanks to the substantial reduction in computational complexity brought by visual token compression. For example, compared to LLaVE-2B based on Aquila-VL, Magic-MM-Embedding-2B reduced the inference latency for MMEB queries from 162.8 ms to 29.9 ms and for VisDoc candidates from 233.6 ms to 57.3 ms. We observed that for query inference latency in VisDoc, Magic-MM-Embedding (2B/8B) had slightly higher latency than GME (2B/8B). This is because, for T$\rightarrow$VD tasks, the system prompt length of InternVL3 is slightly longer than that of Qwen2-VL, given the nearly identical parameter scale of the language models. However, this latency difference can be mitigated in actual deployment using prefix cache techniques~\cite{kwon2023efficient}. We also conducted a comparison with the vanilla InternVL3 architecture. The only difference in Magic-MM-Embedding is the introduction of a parameter-free visual token compression module. We find that, compared with the vanilla architecture, reducing the number of visual tokens by 75\% leads to a significant improvement in inference efficiency. We also provide efficiency comparisons under concurrent requests with vLLM~\cite{kwon2023efficient} deployment in Appendix B.5.

\subsection{Ablation Study}

\noindent\textbf{Ablation Study on the Number of Visual Tokens ($v$).} For a single image tile, we set the number of visual tokens $v$ input to the LLM in InternVL3‑VTC‑2B to 36, 64, 100, and 144 to conduct experiments. All models perform stage 1 generative recovery and stage 2 contrastive warm‑up on 48$\times$NVIDIA A800 (80GB) GPUs,

\begin{wrapfigure}[10]{r}{0.43\linewidth}
  \centering
  \includegraphics[width=\linewidth]{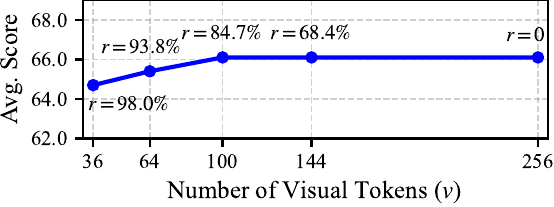}
  \caption{Sensitivity of \#visual tokens.}
  \label{fig:vt_num_ablation}
\end{wrapfigure}

\noindent  using the same full dataset as the 64-visual‑token configuration. We also compare with vanilla InternVL3‑2B ($v=256$), trained with stage 2 contrastive warm‑up. We show the average score on MMEB and VisDoc (The scores of MMEB and VisDoc are in Appendix B.4.), and the attention computation reduction ratio $r = 1 - \left(\frac{v}{256}\right)^2$ on~\cref{fig:vt_num_ablation}. With 64 tokens, the model shows only a 0.7 average score drop compared to the vanilla setting (65.4 vs. 66.1) while reducing attention computation by 93.8\%. Fewer tokens yield limited computation savings but larger performance loss, so we adopt 64 tokens.

\begin{table}[H]
    \centering
    \caption{Ablation study on progressive training pipeline \& reranker.}
    \label{tab:ablation_training_pipeline}
    \setlength\tabcolsep{6pt}
    \resizebox{0.56\linewidth}{!}{
        \begin{tabular}{cccccc}
            \toprule
            \textbf{Stage 1} & \textbf{Stage 2} & \textbf{Stage 3} & \textbf{Reranker} & \textbf{MMEB} & \textbf{VisDoc} \\
            \midrule
            \ding{51} & \ding{55} & \ding{55} & \ding{55} & 1.9 & 0.5 \\
            \ding{55} & \ding{51} & \ding{55} & \ding{55} & 64.7 & 69.8 \\
            \ding{51} & \ding{51} & \ding{55} & \ding{55} & 65.4 & 70.7 \\
            \ding{51} & \ding{55} & \ding{51} & \ding{55} & 67.1 & 70.9 \\
            \rowcolor{brown!10}\ding{51} & \ding{51} & \ding{51} & \ding{55} & 68.0 & 72.1 \\
            \rowcolor{brown!20}\ding{51} & \ding{51} & \ding{51} & \ding{51} & \textbf{70.2} & \textbf{73.3} \\
            \bottomrule
        \end{tabular}
    }
\end{table}

\noindent\textbf{Ablation Study on Progressive Training Pipeline \& Reranker.} We analyze the contribution of each component in our progressive coarse-to-fine training pipeline. The results are reported in~\cref{tab:ablation_training_pipeline}. All experiments are conducted with Magic-MM-Embedding-2B. The results show that training with only stage 1 leads to extremely poor performance (MMEB 1.9, VisDoc 0.5), indicating its insufficiency when used alone. Skipping stage 1 and directly applying stage 2 also results in performance degradation (e.g., MMEB drops from 65.4 to 64.7), showing that stage 1 provides essential multimodal understanding capabilities. In the subsequent contrastive learning training, removing either stage 2 or stage 3 consistently harms performance, while the full three-stage training achieves the best results without reranking (MMEB 68.0, VisDoc 72.1). Finally, introducing a reranker at inference time further improves performance to 70.2 on MMEB and 73.3 on VisDoc, confirming that incorporating reranking into the inference pipeline can further enhance retrieval performance.

\begin{wrapfigure}[21]{r}{0.4\linewidth}
  \centering
  \includegraphics[width=\linewidth]{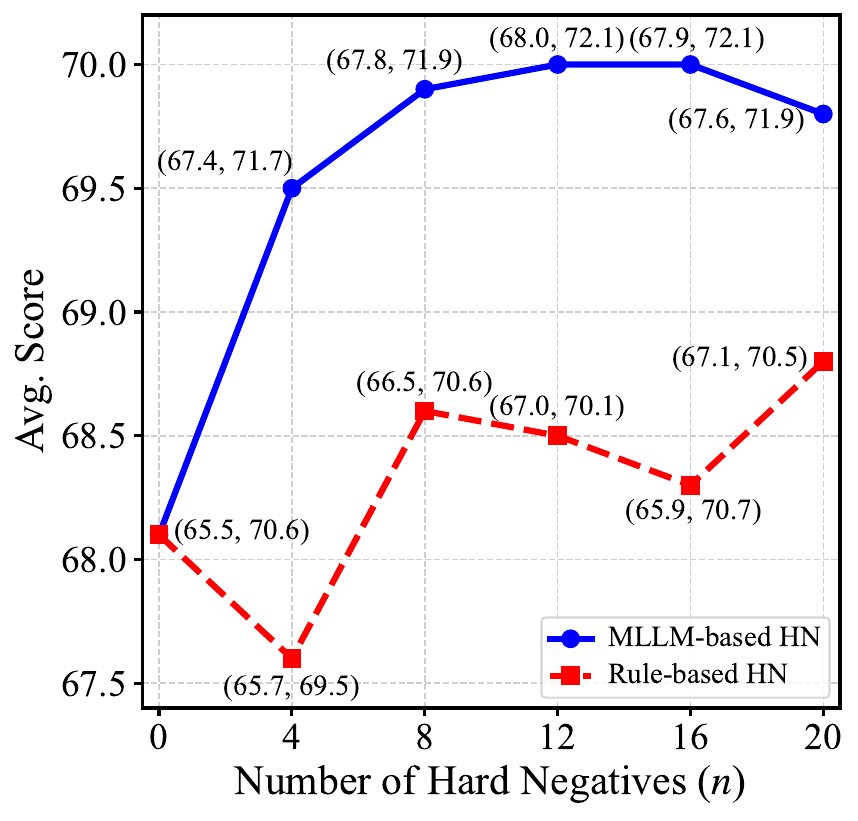}
  \caption{Impact of the number and types of Hard Negatives (HN). We report the average scores of MMEB and VisDoc. The scores in parentheses correspond to MMEB and VisDoc, respectively.}
  \label{fig:ablation_hn}
\end{wrapfigure}

\noindent\textbf{Ablation Study on the Number and Types of Hard Negatives (HN).} We investigate the sensitivity of our model to the number of hard negatives $n$ during stage 3 training, as shown in~\cref{fig:ablation_hn}. All experiments are conducted on Magic-MM-Embedding-2B, where the value of  $n$ is set to 0, 4, 8, 12, 16, and 20. The results show that, compared with using only the standard in-batch negatives sampling strategy, introducing any number of MLLM-based hard negatives consistently yields substantial performance improvements. As $n$ increases, the difficulty of the dataset increases, and the model performance first improves and then slightly declines. For example, on MMEB, the performance peaks at $n = 16$, and further increasing $n$ leads to a mild degradation. To verify the effectiveness of using an MLLM as an ``expert'' to mine hard negatives, we further compare our approach with a rule-based hard negative mining strategy, as shown in~\cref{fig:ablation_hn}. This strategy removes the ground-truth sample from the retrieved top-$K$ candidates and treats the remaining samples as hard negatives, inevitably introducing false negatives. The experimental results show that, for the same value of $n$, models trained with MLLM-based hard negatives consistently and significantly outperform those trained with the same number of rule-based hard negatives.

\noindent\textbf{Ablation Study on Judge Models in Stage 3.} We replace the judge model Qwen3‑VL‑8B~\cite{bai2025qwen3} used in stage 3 training with InternVL3‑8B~\cite{zhu2025internvl3} and retrain Magic‑MM‑Embedding‑2B under the same settings. \cref{tab:ablation_judge_model} shows that the performance with InternVL3‑8B is nearly identical to Qwen3‑VL‑8B, both outperforming the rule‑based baseline. Moreover, the consistency between the two judges across 21 datasets used in stage 3 (2.8M pairs) reaches 80.4\%, indicating that our method is robust to the judge model choice.

\begin{table}[!htbp]
\centering

\begin{minipage}{0.42\linewidth}
  \centering
  \setlength\tabcolsep{10pt}
  \caption{Ablation study of judge models in stage 3.}
  \label{tab:ablation_judge_model}
  \resizebox{\linewidth}{!}{
    \begin{tabular}{lccc}
    \toprule
    \textbf{Judge Model} & \textbf{MMEB} & \textbf{VisDoc} & \textbf{Avg.}\\
    \midrule
    Rule-based & 67.0 & 70.1 & 68.5 \\
    InternVL3-8B & 67.8 & 72.1 & 69.9 \\
    \rowcolor{brown!20} Qwen3-VL-8B & 68.0 & 72.1 & \textbf{70.0} \\
    \bottomrule
    \end{tabular}
  }
\end{minipage}
\hfill
\begin{minipage}{0.52\linewidth}
  \centering
  \setlength\tabcolsep{4pt}
  \caption{Ablation study of visual token compression for training efficiency.}
  \label{tab:vtc_train_efficiency}
  \resizebox{\linewidth}{!}{
  {
  \renewcommand{\arraystretch}{1.24}
    \begin{tabular}{lcccc}
      \toprule
      \textbf{Backbone} & \textbf{Training Duration} & \textbf{MMEB} & \textbf{VisDoc}\\
      \midrule
      InternVL3-2B & 52h 43m 35s & 63.7 & 68.5\\
      \rowcolor{brown!20} InternVL3-VTC-2B & \textbf{24h 38m 28s} & 63.2 & 67.6\\
      \bottomrule
    \end{tabular}
  }
  }
\end{minipage}

\end{table}

\noindent\textbf{Ablation Study on Visual Token Compression for Training Efficiency.} We use InternVL3-VTC-2B and the vanilla InternVL3-2B as backbones to investigate the impact of using vs. not using a token compression module on training efficiency. Both models are trained with contrastive learning on the 16M dataset used during the stage 2 warm-up phase on 48$\times$NVIDIA A800 (80GB) GPUs. During training, the global batch size of the vanilla InternVL3-2B was set to a maximum of 3,456 to ensure sufficient training of the model. To keep the settings aligned, the global batch size of InternVL3-VTC-2B was also set to 3,456. All other training hyperparameters are kept identical. The experimental results are shown in~\cref{tab:vtc_train_efficiency}. We observe that, with almost no degradation in model performance, the proposed visual token compression method can significantly improve training efficiency. For example, for the 2B-scale models, the training time for 2 epochs is reduced from approximately 53 hours to 25 hours.  
\section{Conclusion}

In this work, we identified a critical computational bottleneck in current MLLM-based universal embedding models: \textit{the prohibitively high cost of processing redundant visual tokens}. To address this, we propose the Magic-MM-Embedding series, which uses merely 25\% of the baseline visual tokens, significantly reducing inference latency and training time. Crucially, we demonstrate that when combined with our proposed training strategy, this simplified architecture not only avoids performance degradation but also achieves a new state-of-the-art. Specifically, we introduce a novel three-stage progressive training pipeline: advancing from generative restoration to contrastive self-mining, and finally to task-aware refinement guided by an MLLM as a judge. This strategy progressively enhances the discriminative capability of the embedders in a coarse-to-fine manner. Furthermore, by equipping our proposed efficient embedders with the synergistically trained rerankers, we establish a comprehensive retrieval pipeline. Experimental results demonstrate that our ``Embedder + Reranker'' setup outperforms baselines under the same settings, as well as competitors trained on much larger proprietary datasets, thereby validating the effectiveness of this synergistic inference strategy.

\section*{Acknowledgements}

We thank our former colleague, Yue Sun, for contributions to the data construction in the early stage of the project. We also thank Yaqi Luo (intern) for her contributions to the experimental validation.

\bibliographystyle{splncs04}
\bibliography{main}

\FloatBarrier
\clearpage
\appendix
\title{Supplementary Materials for \\``Magic-MM-Embedding: Towards Visual-Token-Efficient Universal Multimodal Embedding with MLLMs''} 
% \title{Towards Token-Efficient Universal Multimodal Embedding with MLLMs} 

% TODO REVIEW: If the paper title is too long for the running head, you can set
% an abbreviated paper title here. If not, comment out.
\titlerunning{Magic-MM-Embedding}

% TODO FINAL: Replace with your author list. 
% Include the authors' OCRID for the camera-ready version, if at all possible.
% \author{First Author\inst{1}\orcidlink{0000-1111-2222-3333} \and
% Second Author\inst{2,3}\orcidlink{1111-2222-3333-4444} \and
% Third Author\inst{3}\orcidlink{2222--3333-4444-5555}}

\author{Qi Li \and
Yanzhe Zhao \and
Yongxin Zhou \and
Yameng Wang \and
Yandong Yang \and \\
Yuanjia Zhou \and
Jinxiang Liu\thanks{Corresponding author.}}

% TODO FINAL: Replace with an abbreviated list of authors.
\authorrunning{Q.~Li et al.}
% First names are abbreviated in the running head.
% If there are more than two authors, 'et al.' is used.

% TODO FINAL: Replace with your institution list.
\institute{Honor Device Co., Ltd., China\\
% \email{lncs@springer.com}\\
% \url{http://www.springer.com/gp/computer-science/lncs} \and
% ABC Institute, Rupert-Karls-University Heidelberg, Heidelberg, Germany\\
\email{\{liqi20, zhaoyanzhe2, zhouyongxin, wangyameng, yangyandong,\\
zhouyuanjia2, liujinxiang\}@honor.com}
}

\maketitle

\setcounter{figure}{0}
\setcounter{table}{0}

\renewcommand{\thefigure}{S\arabic{figure}}
\renewcommand{\thetable}{S\arabic{table}}

\section{Details of the Visual Token Compression Module}\label{sec:detail_vtc_module}

\begin{figure}[H]
  \centering
  \includegraphics[width=\linewidth]{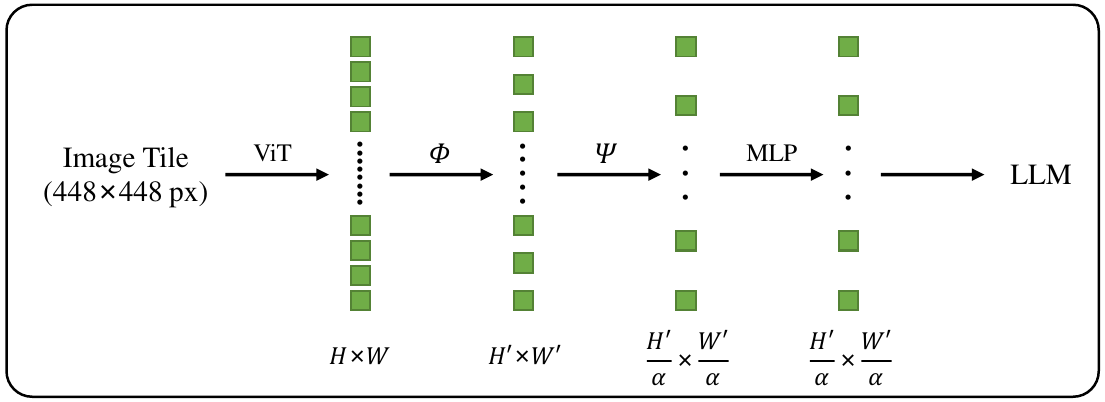}
  \caption{Visual token compression workflow. $\Phi$ denotes the bilinear interpolation operation and $\Psi$ denotes the pixel unshuffle operation. The $H \times W$ visual tokens generated by the ViT are downsampled twice to produce $\frac{H'}{\alpha} \times \frac{W'}{\alpha}$ visual tokens, which are finally projected through an MLP and fed into the LLM.}
  \label{fig:vtc_module}
\end{figure}

We use the official PyTorch function \texttt{torch.nn.functional.interpolate} to implement bilinear interpolation. For the pixel unshuffle operation, we adopt the official implementation from InternVL3~\cite{zhu2025internvl3}, and set the downscaling factor $\alpha$ to 2, meaning that the pixel unshuffle operation reduces the number of visual tokens to one quarter of the input. The overall process of visual token compression is illustrated in~\cref{fig:vtc_module}. Each image tile is first processed by the ViT to produce $H \times W$ visual tokens. After two downsampling stages using bilinear interpolation and pixel unshuffle, $\frac{H'}{\alpha} \times \frac{W'}{\alpha}$ visual tokens are obtained, which are finally projected into the LLM through an MLP.

\section{Extra Experiments}\label{sec:extra_exp}

\subsection{Ablation Study of Different Visual Token Compression Methods}\label{subsec:ablation_vtc_method}

\noindent We compare three visual token compression (VTC) methods on the 2B model: 2D convolution (Conv2d, learnable, 9.4M parameters), 2D adaptive average pooling (AdaptiveAvgPool2d, parameter‑free, direct averaging), and bilinear interpolation (ours, parameter‑free, preserves locality). Each module is concatenated with pixel unshuffle to feed 64 visual tokens to the LLM for each image tile. All three models are trained using 1.6M samples from stage 1 generative recovery, followed by 1M multimodal pairs without negatives sampled from stage 2 data.~\cref{tab:vtc_method} shows that bilinear interpolation achieves the highest score. 2D adaptive average pooling loses fine details, hurting the T$\rightarrow$VD task. 2D convolution learns from scratch, requiring mutual adaptation between the MLLM and the module, which leads to collapse. Bilinear interpolation is learning-free, avoids adaptation cost, and retains local details, thus outperforming both.

\begin{table}[!htbp]
    \centering
    \setlength\tabcolsep{10pt}
    \caption{Ablation analysis of VTC methods.}
    \label{tab:vtc_method}
    \resizebox{0.5\linewidth}{!}{
    \begin{tabular}{lccc}
    \toprule
    \textbf{VTC Method} & \textbf{MMEB} & \textbf{VisDoc} & \textbf{Avg.}\\
    \midrule
    Conv2d & 35.1 & 0.7 & 17.9 \\
    AdaptiveAvgPool2d & 57.4 & 63.8 & 60.6 \\
    \rowcolor{brown!20}Bilinear Interpolation & 57.4 & 64.6 & \textbf{61.0} \\
    \bottomrule
    \end{tabular}
    }
\end{table}

\subsection{Ablation Study of Multimodal Understanding Ability vs. Number of Visual Tokens}\label{subsec:ablation_mme_vs_v_num}
We investigate the impact of the number of visual tokens fed into the LLM on multimodal understanding performance. The results are summarized in~\cref{tab:vtc_model_on_mme}. Specifically, we set the number of visual tokens  generated by a single image tile in InternVL3‑VTC‑2B to 36, 64, 100, and 144. All models are trained using 32M samples from the stage 1 training. We further compare these visual‑token‑compressed models with the vanilla InternVL3‑2B. All evaluations are conducted on the MME benchmark~\cite{fu2023mme}, where we report accuracy scores. Performance is measured across 14 subtasks, including existence, count, position, color, poster, celebrity, scene, landmark, artwork, OCR, commonsense reasoning, numerical calculation, text translation, and code reasoning. We also report the average score over all tasks. Our results show that all models with visual token compression achieve similar average scores (around 85.5), and outperform the vanilla InternVL3‑2B. This indicates that models with visual token compression after the stage 1 generative recovery can effectively preserve multimodal understanding capability.
\begin{table}[H]
    \centering
    \setlength\tabcolsep{6pt}
    \caption{Ablation results on multimodal understanding ability vs. number of visual tokens. \#$VT$ denotes the number of visual tokens fed into the LLM.}
    \label{tab:vtc_model_on_mme}
    \resizebox{\linewidth}{!}{
        \begin{tabular}{lrcccccccc}
            \toprule
            \textbf{Model} & \textbf{\#$VT$} & \textbf{Existence} & \textbf{Count} & \textbf{Position} & \textbf{Color} & \textbf{Posters} & \textbf{Celebrity} & \textbf{Scene} & \textbf{Landmark}\\
            \midrule
            InternVL3-VTC & 36 & 98.3 & 91.7 & 85.0 & 93.3 & 83.0 & 90.6 & 89.0 & 87.3 \\
            \rowcolor{brown!20} InternVL3-VTC & 64 & 98.3 & 90.0 & 81.7 & 95.0 & 85.0 & 88.5 & 87.5 & 86.8 \\
            InternVL3-VTC & 100 & 98.3 & 88.3 & 81.7 & 95.0 & 84.4 & 90.5 & 85.8 & 87.8 \\
            InternVL3-VTC & 144 & 98.3 & 90.0 & 80.0 & 95.0 & 83.3 & 89.4 & 88.8 & 88.5 \\
            InternVL3 (Vanilla) & 256 & 100.0 & 83.3 & 75.0 & 91.7 & 87.4 & 86.5 & 85.5 & 89.0 \\

            \midrule
            \textbf{Model} & \textbf{\#$VT$} & \textbf{Artwork} & \textbf{OCR} & \textbf{Comm.} & \textbf{Num.} & \textbf{Text.} & \textbf{Code.} & \multicolumn{2}{c}{\textbf{\textit{Average Score}}}\\
            \midrule
            InternVL3-VTC & 36 & 82.5 & 85.0 & 75.0 & 77.5 & 95.0 & 70.0 & \multicolumn{2}{c}{85.9} \\
            \rowcolor{brown!20} InternVL3-VTC & 64 & 83.8 & 87.5 & 78.6 & 72.5 & 95.0 & 75.0 & \multicolumn{2}{c}{\textbf{86.1}} \\
            InternVL3-VTC & 100 & 83.8 & 85.0 & 77.9 & 67.5 & 97.5 & 72.5 & \multicolumn{2}{c}{85.4} \\
            InternVL3-VTC & 144 &  84.5 & 85.0 & 74.3 & 70.0 & 97.5 & 72.5  & \multicolumn{2}{c}{85.5} \\
            InternVL3 (Vanilla) & 256 & 85.5 & 90.0 & 67.9 & 70.0 & 97.5 & 77.5 & \multicolumn{2}{c}{84.9} \\
            \bottomrule
        \end{tabular}
    }
\end{table}

\subsection{Ablation Study on LoRA Rank}\label{subsec:ablation_lora}
\cref{tab:lora_rank} presents the ablation results of LoRA rank. We conducted experiments using Magic‑MM‑Embedding‑2B during the stage 2 contrastive learning warm‑up phase, using a global batch size of 6,144. We set the LoRA rank to 8, 16, and 32 for the experiments. We found that when the LoRA rank is set to 16, the average score on MMEB and VisDoc is optimal. Further increasing the LoRA rank leads to a decline in overall performance. Therefore, in all training of the embedder, the LoRA rank is set to 16.

\begin{table}[H]
    \centering
    \setlength\tabcolsep{10pt}
    \caption{Ablation analysis of LoRA rank.}
    \label{tab:lora_rank}
    \resizebox{0.46\linewidth}{!}{
        \begin{tabular}{cccc}
            \toprule
            \textbf{LoRA Rank} & \textbf{MMEB} & \textbf{VisDoc} & \textbf{Avg.}\\
            \midrule
            8 & 62.9 & 68.0 & 65.5 \\
            \rowcolor{brown!20} 16 & 62.9 & 68.4 & \textbf{65.7} \\
            32 & 62.6 & 67.6 & 65.1 \\
            \bottomrule
        \end{tabular}
    }
\end{table}

\subsection{Details of Embedding Performance vs. Number of Visual Tokens}\label{subsec:embed_perf_ablation_vt_num}

For a single image tile, we vary the number of visual tokens $v$ fed into the LLM in InternVL3‑VTC‑2B (36, 64, 100, and 144). All variants are trained with stage 1 generative recovery and stage 2 contrastive warm‑up on 48$\times$NVIDIA A800 (80GB) GPUs, using a global batch size of 3,456 and the same full dataset as the 64‑token configuration. We also include the vanilla InternVL3‑2B ($v=256$), trained with stage 2 contrastive warm‑up, for comparison.~\cref{tab:vt_num_vs_mmeb_visdoc} shows the performance on MMEB and VisDoc with different numbers of visual tokens, along with the attention computation reduction ratio $r = 1 - \left(\frac{v}{256}\right)^2$. Compared with the vanilla setting, using 64 visual tokens reduces computation by 93.8\%, with only a 0.7 drop in average score.

\begin{table}[!htbp]
    \centering
    \setlength\tabcolsep{10pt}
    \caption{Performance vs. number of visual tokens. \#$VT$ denotes the number of visual tokens fed into the LLM.}
    \label{tab:vt_num_vs_mmeb_visdoc}
    \resizebox{0.52\linewidth}{!}{
    \begin{tabular}{ccccc}
    \toprule
    \textbf{\#$VT$} & \textbf{MMEB} & \textbf{VisDoc} & \textbf{Avg.} & \textbf{$r$(\%)}\\
    \midrule
    36 & 62.6 & 66.8 & 64.7 & 98.0\\
    \rowcolor{brown!20} 64 & 63.2 & 67.6 & 65.4 & \textbf{93.8}\\
    100 & 63.3 & 68.9 & 66.1 & 84.7\\
    144 & 63.1 & 69.1 & 66.1 & 68.4\\
    256 (Vanilla) & 63.7 & 68.5 & 66.1 & 0\\
    \bottomrule
    \end{tabular}
    }
\end{table}

\subsection{Efficiency Comparison under Concurrent Requests with vLLM Deployment}\label{subsec:vllm_efficiency}
We deploy embedders with vLLM~\cite{kwon2023efficient} on an NVIDIA L20 (48GB) GPU and measure average latency and throughput under concurrent requests on both MMEB and VisDoc. 5,000 queries and candidates are randomly sampled from the MMEB and VisDoc training sets. To ensure fairness, natural images were resized to 448$\times$448, and visual document images were resized to 896$\times$896. We select VLM2Vec~\cite{VLM2Vec} (backbone: Qwen2-VL) and vanilla InternVL3~\cite{zhu2025internvl3} as baselines and compare them under settings where the number of concurrent requests used in a single run is 5, 10, and 20. The comparison results are shown in~\cref{tab:efficiency_comparison_with_vllm}. Except for the VisDoc queries, where the three models show nearly identical metrics (as these queries are pure text and do not involve visual tokens), our model achieves significantly higher efficiency than the baseline models across all metrics.

\begin{table}[H]
    \centering
    \setlength\tabcolsep{4pt}
    \caption{Inference efficiency comparison after deployment with vLLM. $NCR$ denotes the number of concurrent requests used in a single run. \textbf{$l_q$} and \textbf{$l_c$} mean the average latency (millisecond) of query inference and candidate inference, respectively. $TP$ represents the throughput (req/s) of the model in processing requests.}
    \label{tab:efficiency_comparison_with_vllm}
    \resizebox{\linewidth}{!}{
        \begin{tabular}{>{\centering\arraybackslash}p{1.8cm}lrc>{\centering\arraybackslash}p{1.4cm}c>{\centering\arraybackslash}p{1.4cm}c>{\centering\arraybackslash}p{1.4cm}c>{\centering\arraybackslash}p{1.4cm}}
            \toprule
            \multirow{2}{*}{$NCR$} & \multirow{2}{*}{\textbf{Model}} & \multirow{2}{*}{\textbf{Backbone (Model Size)}} & \multicolumn{4}{c}{\textbf{MMEB}} & \multicolumn{4}{c}{\textbf{VisDoc}} \\
            \cmidrule(lr){4-7} \cmidrule(lr){8-11}
            & & & \textbf{$l_q$ $(\downarrow)$} & \textbf{$TP$ $(\uparrow)$} & \textbf{$l_c$ $(\downarrow)$} & \textbf{$TP$ $(\uparrow)$} & \textbf{$l_q$ $(\downarrow)$} & \textbf{$TP$ $(\uparrow)$} & \textbf{$l_c$ $(\downarrow)$} & \textbf{$TP$ $(\uparrow)$} \\
            \midrule
            \multirow{3}{*}{5} & VLM2Vec & Qwen2-VL (2.2B) & 144.1 & 34.7 & 85.6 & 58.4 & 21.3 & 234.7 & 633.3 & 7.9 \\
             & InternVL3 & InternVL3 (1.9B) & 98.4 & 50.8 & 64.6 & 77.4 & 20.8 & 240.4 & 378.8 & 13.2 \\
             & \cellcolor{brown!20}\textbf{Magic-MM-Embedding} & \cellcolor{brown!20}InternVL3-VTC (1.9B) & \cellcolor{brown!20}\textbf{69.3} & \cellcolor{brown!20}\textbf{72.2} & \cellcolor{brown!20}\textbf{47.6} & \cellcolor{brown!20}\textbf{105.0} & \cellcolor{brown!20}\textbf{20.6} & \cellcolor{brown!20}\textbf{242.7} & \cellcolor{brown!20}\textbf{205.4} & \cellcolor{brown!20}\textbf{24.3} \\
                                                                                     \midrule
            \multirow{3}{*}{10} & VLM2Vec & Qwen2-VL (2.2B) & 272.3 & 36.7 & 152.7 & 65.5 & \textbf{31.9} & \textbf{313.5} & 1266.7 & 7.9 \\
             & InternVL3 & InternVL3 (1.9B) & 180.1 & 55.5 & 108.7 & 92.0 & 32.4 & 308.6 & 772.3 & 12.9 \\
             & \cellcolor{brown!20}\textbf{Magic-MM-Embedding} & \cellcolor{brown!20}InternVL3-VTC (1.9B) & \cellcolor{brown!20}\textbf{110.9} & \cellcolor{brown!20}\textbf{90.2} & \cellcolor{brown!20}\textbf{71.3} & \cellcolor{brown!20}\textbf{140.3} & \cellcolor{brown!20}32.7 & \cellcolor{brown!20}305.8 & \cellcolor{brown!20}\textbf{385.6} & \cellcolor{brown!20}\textbf{25.9} \\
                                                                                                 \midrule
            \multirow{3}{*}{20} & VLM2Vec & Qwen2-VL (2.2B) & 555.4 & 36.0 & 305.8 & 65.4 & \textbf{45.2} & \textbf{442.5} & 2548.0 & 7.8 \\
             & InternVL3 & InternVL3 (1.9B) & 342.2 & 58.4 & 193.6 & 103.3 & 49.9 & 400.8 & 1544.1 & 13.0 \\
             & \cellcolor{brown!20}\textbf{Magic-MM-Embedding} & \cellcolor{brown!20}InternVL3-VTC (1.9B) & \cellcolor{brown!20}\textbf{200.8} & \cellcolor{brown!20}\textbf{99.6} & \cellcolor{brown!20}\textbf{131.4} & \cellcolor{brown!20}\textbf{152.2} & \cellcolor{brown!20}48.7 & \cellcolor{brown!20}410.7 & \cellcolor{brown!20}\textbf{802.4} & \cellcolor{brown!20}\textbf{24.9} \\
            \bottomrule
        \end{tabular}
    }
\end{table}

\section{Input Templates}\label{sec:input_temp}
We list below the input templates used in the implementation of the embedding models and the reranking models.

\subsection{Input Templates Used for Implementing Embedding Models}\label{subsec:input_temp_for_embed}
\begin{figure}[H]
  \centering
  \includegraphics[width=\linewidth]{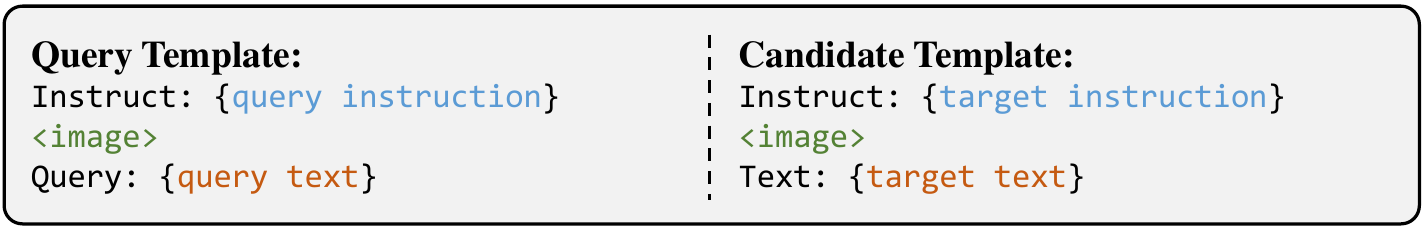}
  \caption{Templates for queries and candidates used in the embedding model.}
  \label{fig:temp_que_can_in_embedding}
\end{figure}

\noindent \cref{fig:temp_que_can_in_embedding} shows the query and candidate templates used in the embedding model.

\begin{figure}[H]
  \centering
  \includegraphics[width=\linewidth]{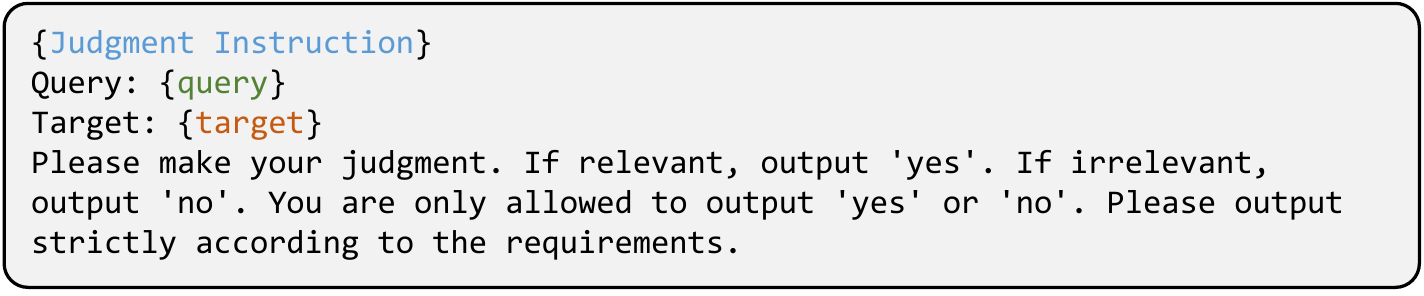}
  \caption{Template in stage 3 for the MLLM expert to judge query-candidate relevance.}
  \label{fig:temp_stage3_mllm_judge}
\end{figure}

\noindent \cref{fig:temp_stage3_mllm_judge} presents the template used in stage 3 for the MLLM expert to judge the relevance of query-candidate pairs.

\subsection{Input Templates Used for Implementing Reranking Models}\label{subsec:input_temp_for_reranking}

\begin{figure}[H]
  \centering
  \includegraphics[width=\linewidth]{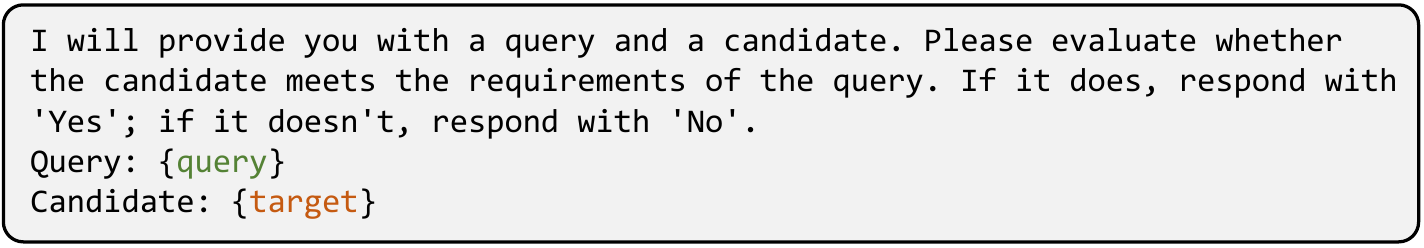}
  \caption{Template for pointwise reranking.}
  \label{fig:temp_pointwise_reranking}
\end{figure}

\begin{figure}[H]
  \centering
  \includegraphics[width=\linewidth]{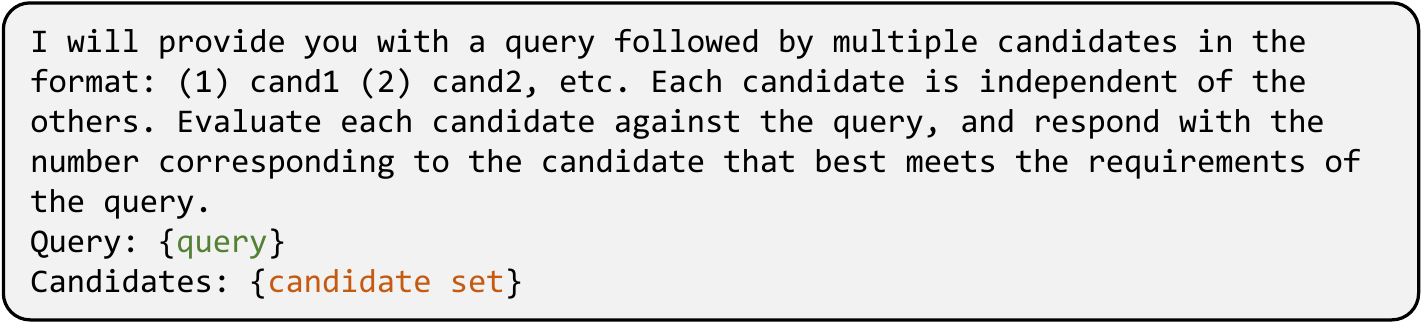}
  \caption{Template for listwise reranking.}
  \label{fig:temp_listwise_reranking}
\end{figure}

\noindent The template for pointwise reranking is shown in~\cref{fig:temp_pointwise_reranking}, and the template for listwise reranking is shown in~\cref{fig:temp_listwise_reranking}.

\newpage

\section{Task Instructions for Embedding}\label{sec:task_inst_for_embed}
The task instructions for embedding across different datasets are presented in~\cref{tab:embedding_instruction_1,tab:embedding_instruction_2,tab:embedding_instruction_3}.
\begin{table}[H]
\centering
\scriptsize
\caption{Query and target instructions for different datasets (Part 1 of 3). For the queries in mmE5-synthetic~\cite{chen2025mme5}, we use the original instructions from the dataset.}
\resizebox{\linewidth}{!}{
\setlength{\tabcolsep}{4pt}
\begin{tabular}{llm{6cm}m{3.5cm}}
\toprule
\bf Task& \bf Dataset & \bf Query Instruction & \bf Target Instruction \\
\midrule

T$\rightarrow$T
&BAAI-MTP~\cite{baai-mtp} & Retrieve relevant texts based on a given query. & Represent the given text. \\
\midrule

\multirow{2}{*}{I$\rightarrow$I}
&ImageNet-1K~\cite{deng2009imagenet} & Find a image that looks similar to the provided image. & Represent the given image. \\ \cmidrule(lr){2-4}
&NIGHTS~\cite{fu2023dreamsim} & Find a day-to-day image that looks similar to the provided image. & Represent the given image. \\
\midrule

\multirow{8}{*}{T$\rightarrow$I}
&\makecell[l]{BLIP Bootstrapped\\Image-Text Pairs~\cite{li2022blip}}  & Retrieve relevant images based on a given query. & Represent the given image. \\ \cmidrule(lr){2-4}
&VisDial~\cite{das2017visual}  & Represent the given dialogue about an image, which is used for image retrieval. & Represent the given image. \\ \cmidrule(lr){2-4}
&VisualNews~\cite{liu2021visual}  & Retrieve an image of the given query. & Represent the given image. \\ \cmidrule(lr){2-4}
&MSCOCO~\cite{lin2014microsoft}  & Find me an everyday image that matches the given query. & Represent the given image. \\ \cmidrule(lr){2-4}
&Flickr30K~\cite{plummer2015flickr30k}  & Find me an everyday image that matches the given query. & Represent the given image. \\ \cmidrule(lr){2-4}
&ShareGPT4V~\cite{chen2024sharegpt4v}  & Find me an everyday image that matches the given query. & Represent the given image. \\ \cmidrule(lr){2-4}
&Urban1K~\cite{longclip}  & Find me an everyday image that matches the given query. & Represent the given image. \\ \cmidrule(lr){2-4}
&mmE5-synthetic~\cite{chen2025mme5}  & - & Represent the given image. \\
\midrule

\multirow{6}{*}{T$\rightarrow$VD}
&Docmatix~\cite{laurenccon2024building}  & Retrieve relevant visual documents based on a given query. & \makecell[l]{Represent the given visual\\documents.} \\ \cmidrule(lr){2-4}
&Colpali~\cite{faysse2024colpali}  & Retrieve relevant visual documents based on a given query. & \makecell[l]{Represent the given visual\\documents.} \\ \cmidrule(lr){2-4}
&VisRAG~\cite{VisRAG}  & Retrieve relevant visual documents based on a given query. & \makecell[l]{Represent the given visual\\documents.} \\ \cmidrule(lr){2-4}
&ViDoSeek~\cite{wang2025vidorag}  & Retrieve relevant visual documents based on a given query. & \makecell[l]{Represent the given visual\\documents.} \\ \cmidrule(lr){2-4}
&MMLongBench~\cite{ma2024mmlongbench}  & Retrieve relevant visual documents based on a given query. & \makecell[l]{Represent the given visual\\documents.} \\ \cmidrule(lr){2-4}
&Wiki-SS-NQ~\cite{ma2024unifying}  & Find the document image that can answer the given query. & \makecell[l]{Represent the given visual\\documents.} \\
\bottomrule
\end{tabular}}
\label{tab:embedding_instruction_1}
\end{table}

\newpage

\begin{table}[H]
\centering
\scriptsize
\caption{Query and target instructions for different datasets (Part 2 of 3). For the queries in mmE5-synthetic~\cite{chen2025mme5}, we use the original instructions from the dataset.}
\resizebox{\linewidth}{!}{
\setlength{\tabcolsep}{4pt}
\begin{tabular}{llm{6cm}m{3.5cm}}
\toprule
\bf Task& \bf Dataset & \bf Query Instruction & \bf Target Instruction \\
\midrule

\multirow{16}{*}{I$\rightarrow$T}
&ImageNet-1K~\cite{deng2009imagenet}  & Represent the given image for classification. & Represent the given text. \\ \cmidrule(lr){2-4}
&HatefulMemes~\cite{kiela2020hateful}  & Represent the given image for binary classification to determine whether it constitutes hateful speech or not. & Represent the given text. \\ \cmidrule(lr){2-4}
&VOC2007~\cite{everingham2015pascal}  & Identify the object shown in the image. & Represent the given text. \\ \cmidrule(lr){2-4}
&SUN397~\cite{xiao2010sun}  & Identify the scene shown in the image. & Represent the given text. \\ \cmidrule(lr){2-4}
&Place365~\cite{zhou2017places}  & Identify the scene shown in the image. & Represent the given text. \\ \cmidrule(lr){2-4}
&ImageNet-A~\cite{hendrycks2021natural}  & Represent the given image for classification. & Represent the given text \\ \cmidrule(lr){2-4}
&ImageNet-R~\cite{hendrycks2021many}  & Represent the given image for classification. & Represent the given text \\ \cmidrule(lr){2-4}
&ObjectNet~\cite{barbu2019objectnet}  & Identify the object shown in the image. & Represent the given text \\ \cmidrule(lr){2-4}
&Country-211~\cite{clip2021Radford}  & Identify the country depicted in the image. & Represent the given text \\ \cmidrule(lr){2-4}
&VisualNews~\cite{liu2021visual}  & Find a caption for the news in the given photo. & Represent the given text. \\ \cmidrule(lr){2-4}
&MSCOCO~\cite{lin2014microsoft}  & Find an image caption describing the given everyday image. & Represent the given text. \\ \cmidrule(lr){2-4}
&Flickr30K~\cite{plummer2015flickr30k}  & Find an image caption describing the given image. & Represent the given text. \\ \cmidrule(lr){2-4}
&ShareGPT4V~\cite{chen2024sharegpt4v}  & Find an image caption describing the given image. & Represent the given text. \\ \cmidrule(lr){2-4}
&Urban1K~\cite{longclip}  & Find an image caption describing the given image. & Represent the given text. \\ \cmidrule(lr){2-4}
&SugarCrepe~\cite{hsieh2023sugarcrepe}  & Find an image caption describing the given image. & Represent the given text. \\ \cmidrule(lr){2-4}
&mmE5-synthetic~\cite{chen2025mme5}  & - & Represent the given text. \\
\midrule

\multirow{12}{*}{IT$\rightarrow$T}
&Docmatix~\cite{laurenccon2024building}  & Represent the given image with the given query and retrieve the answer. & Represent the given text. \\ \cmidrule(lr){2-4}
&OK-VQA~\cite{marino2019ok}  & Represent the given image with the given query and retrieve the answer. & Represent the given text. \\ \cmidrule(lr){2-4}
&A-OKVQA~\cite{schwenk2022okvqa}  & Represent the given image with the given query and retrieve the answer. & Represent the given text. \\ \cmidrule(lr){2-4}
&DocVQA~\cite{mathew2021docvqa}  & Represent the given image with the given query and retrieve the answer. & Represent the given text. \\ \cmidrule(lr){2-4}
&InfographicVQA~\cite{mathew2022infographicvqa}  & Represent the given image with the given query and retrieve the answer. & Represent the given text. \\ \cmidrule(lr){2-4}
&ChartQA~\cite{masry2022chartqa}  & Represent the given image with the given query and retrieve the answer. & Represent the given text. \\ \cmidrule(lr){2-4}
&Visual7W~\cite{zhu2016visual7w}  & Represent the given image with the given query and retrieve the answer. & Represent the given text. \\ \cmidrule(lr){2-4}
&ScienceQA~\cite{lu2022learn}  & Represent the given image with the given query and retrieve the answer. & Represent the given text. \\ \cmidrule(lr){2-4}
&VizWiz~\cite{gurari2018vizwiz}  & Represent the given image with the given query and retrieve the answer. & Represent the given text. \\ \cmidrule(lr){2-4}
&GQA~\cite{hudson2019gqa}  & Represent the given image with the given query and retrieve the answer. & Represent the given text. \\ \cmidrule(lr){2-4}
&TextVQA~\cite{singh2019towards}  & Represent the given image with the given query and retrieve the answer. & Represent the given text. \\ \cmidrule(lr){2-4}
&mmE5-synthetic~\cite{chen2025mme5}  & - & Represent the given text. \\

\bottomrule
\end{tabular}}
\label{tab:embedding_instruction_2}
\end{table}

\newpage

\begin{table}[H]
\centering
\scriptsize
\caption{Query and target instructions for different datasets (Part 3 of 3). For the queries in mmE5-synthetic~\cite{chen2025mme5}, we use the original instructions from the dataset.}
\resizebox{\linewidth}{!}{
\setlength{\tabcolsep}{4pt}
\begin{tabular}{llm{6cm}m{3.5cm}}
\toprule
\bf Task& \bf Dataset & \bf Query Instruction & \bf Target Instruction \\

\midrule
\multirow{3}{*}{T$\rightarrow$IT}
&WebQA~\cite{chang2022webqa}  & Find a related image and text content from Wikipedia that answers the given query. & Represent the given Wikipedia image with related text information. \\ \cmidrule(lr){2-4}
&EDIS~\cite{liu2023edis}  & Find a related image and text content from a news that matches the provided query. & Represent the given image with related text information. \\ \cmidrule(lr){2-4}
&mmE5-synthetic~\cite{chen2025mme5}  & - & Represent the given image with related text information. \\

\midrule
\multirow{8}{*}{IT$\rightarrow$I}
&MegaPairs~\cite{zhou2025megapairs}  & Represent the given image with the given query and retrieve the related images. & Represent the given image. \\ \cmidrule(lr){2-4}
&CIRR~\cite{liu2021image}  & Given an image, find a similar everyday image with the described changes as the given query. & Represent the given image. \\ \cmidrule(lr){2-4}
&N24News~\cite{wang2022n24news}  & Represent the given news image with the given query for domain classification. & Represent the given text. \\ \cmidrule(lr){2-4}
&MSCOCO~\cite{lin2014microsoft}  & Select the portion of the image where the object label is represented by the given query. & Represent the cropped image. \\  \cmidrule(lr){2-4}
&FashionIQ~\cite{wu2021fashion}  & Find an image to match the fashion image and style note. & Represent the given image. \\ \cmidrule(lr){2-4}
&Visual7W-Pointing~\cite{zhu2016visual7w}  & Select the portion of the image that answers the given query. & Represent the cropped image. \\ \cmidrule(lr){2-4}
&RefCOCO~\cite{kazemzadeh2014referitgame}  & Select the portion of the image where the object label is represented by the given query. & Represent the cropped image. \\ \cmidrule(lr){2-4}
&mmE5-synthetic~\cite{chen2025mme5}  & - & Represent the given image. \\
\midrule

\multirow{2}{*}{IT$\rightarrow$IT}
&OVEN~\cite{hu2023open}  & Retrieve a Wikipedia image-description pair that provides evidence for the given query. & Represent the given image with related text information. \\  \cmidrule(lr){2-4}
&RefCOCO-Matching~\cite{kazemzadeh2014referitgame}  & Select the identical object in the image that follows the given query. & Represent the object in the image that follows the given text. \\

\bottomrule
\end{tabular}}
\label{tab:embedding_instruction_3}
\end{table}

\newpage
\section{Instructions for MLLM-as-a-Judge in Stage 3}\label{sec:task_inst_for_mllm_judge_stage3}
During stage 3 training, the instructions for using the MLLM as an expert to judge the relevance of query and candidate pairs across different datasets are presented in~\cref{tab:judge_instruction}.
\begin{table}[H]
\centering
\scriptsize
\caption{Instructions for MLLM judgment in the stage 3. For the HatefulMemes dataset~\cite{kiela2020hateful}, because it has only Yes/No labels, removing the ground truth leaves only correct negative samples. Therefore, we did not use MLLMs to judge this dataset.}
\resizebox{\linewidth}{!}{
\setlength{\tabcolsep}{4pt}
\begin{tabular}{llm{11cm}}
\toprule
\bf Domain & \bf Dataset & \bf Judgment Instruction \\
\midrule

\multirow{19}{*}{MMEB}
&ImageNet-1K~\cite{deng2009imagenet}  & Determine whether a given image contains an object specified by a class label. \\ \cmidrule(lr){2-3}
&N24News~\cite{wang2022n24news}  & Given news containing both image and text content, determine whether the category of the news matches the given category. \\ \cmidrule(lr){2-3}
&VOC2007~\cite{everingham2015pascal}  & Determine whether a given image contains an object specified by a class label. \\ \cmidrule(lr){2-3}
&SUN397~\cite{xiao2010sun}  & Determine whether the given image matches the given scene description. \\ \cmidrule(lr){2-3}

&OK-VQA~\cite{marino2019ok}  & Given a reference image and a question, determine whether the provided answer is correct. \\ \cmidrule(lr){2-3}
&A-OKVQA~\cite{schwenk2022okvqa}  & Given a reference image and a question, determine whether the provided answer is correct. \\ \cmidrule(lr){2-3}
&DocVQA~\cite{mathew2021docvqa}  & Given a reference image and a question, determine whether the provided answer is correct. \\ \cmidrule(lr){2-3}
&InfographicsVQA~\cite{mathew2022infographicvqa}  & Given a reference image and a question, determine whether the provided answer is correct. \\ \cmidrule(lr){2-3}
&ChartQA~\cite{masry2022chartqa}  & Given a reference image and a question, determine whether the provided answer is correct. \\ \cmidrule(lr){2-3}
&Visual7W~\cite{zhu2016visual7w}  & Given a reference image and a question, determine whether the provided answer is correct. \\ \cmidrule(lr){2-3}

&VisDial~\cite{das2017visual}  & Determine whether the given dialogue text is relevant to the given target image. \\ \cmidrule(lr){2-3}
&CIRR~\cite{liu2021image}  & Given a text instruction, a reference image, and a target image, determine whether the reference image transformed by the text instruction is relevant to the target image. \\ \cmidrule(lr){2-3}

&VisualNews (T$\rightarrow$I)~\cite{liu2021visual}  & Determine whether the given query text is relevant to the given image. \\ \cmidrule(lr){2-3}
&VisualNews (I$\rightarrow$T)~\cite{liu2021visual}  & Determine whether the given text can serve as a caption for the given image. \\ \cmidrule(lr){2-3}
&MSCOCO (T$\rightarrow$I)~\cite{lin2014microsoft}  & Determine whether the given query text is relevant to the given image. \\ \cmidrule(lr){2-3}
&MSCOCO (I$\rightarrow$T)~\cite{lin2014microsoft}  & Determine whether the given text can serve as a caption for the given image. \\ \cmidrule(lr){2-3}
&NIGHTS~\cite{fu2023dreamsim}  & Determine whether the two given images are similar. \\ \cmidrule(lr){2-3}
&WebQA~\cite{chang2022webqa}  & Determine whether the query text is relevant to the given image-text mixed content. \\ \cmidrule(lr){2-3}

&MSCOCO (IT$\rightarrow$I)~\cite{lin2014microsoft}  & Given a reference image, a object label expressed in text pointing to an object in the reference image, and a crop extracted from an image, determine whether the text label points to the given crop. \\
\midrule

\multirow{2}{*}{VisDoc}
&Colpali~\cite{faysse2024colpali} & Determine whether the given query text is relevant to the given visual document image. \\ \cmidrule(lr){2-3}
&VisRAG~\cite{VisRAG} & Determine whether the given query text is relevant to the given visual document image. \\
\bottomrule
\end{tabular}}
\label{tab:judge_instruction}
\end{table}
\newpage

\section{Data for Progressive Coarse-to-Fine Training}\label{sec:data_construct}
The training data used in stage 1 are summarized in \cref{tab:stage1_data}.

\begin{table}[H]
\centering
\small
\caption{Details of the training data for restoring multimodal understanding and generation capabilities in stage 1.}
\setlength{\tabcolsep}{2pt}
\resizebox{\linewidth}{!}{
\begin{tabular}{p{5cm}>{\centering\arraybackslash}p{1.8cm}p{9.5cm}}
\toprule
\textbf{Task} & \textbf{\#Samples} & \textbf{Datasets} \\ 
\midrule
\multirow{3}{*}{Multimodal Instruction Data} & \multirow{3}{*}{12.8M} & Infinity-MM~\cite{gu2024infinity}, Bunny-v1.1~\cite{he2024efficient}, VLFeedback~\cite{li2023silkie}, RLHF-V~\cite{yu2023rlhf},  \\
                                        & & RLAIF-V~\cite{yu2024rlaifv}, DT-VQA~\cite{zhang2024exploring}, LLaVA Visual Instruct 150K~\cite{liu2024improved}, \\
                                        & & Monkey~\cite{li2024monkey}, LVIS-Instruct4V~\cite{wang2023see}, LRV-Instruction~\cite{liu2023mitigating} \\ 
\addlinespace[4pt]
\multirow{2}{*}{Pure Text Instruction  Data}                   & \multirow{2}{*}{8.8M}  & Infinity Instruct~\cite{li2025infinity}, ShareGPT-Chinese-English-90k~\cite{sharegpt_chinese_english_90k},  \\
                                        & & firefly-train-1.1M~\cite{Firefly}, COIG-CQIA~\cite{bai2024coig} \\
\addlinespace[4pt]
Captioning                              & 1.7M  & ShareGPT4V~\cite{chen2024sharegpt4v}, \textcolor{blue}{In-house} \\
\addlinespace[4pt]
\multirow{3}{*}{Grounding}              & \multirow{3}{*}{5.7M}  & RefCOCO~\cite{kazemzadeh2014referitgame}, RefCOCO+~\cite{kazemzadeh2014referitgame}, RefCOCOg~\cite{mao2016generation}, Objects365 v2~\cite{shao2019objects365}, \\
                                        & & Visual Genome~\cite{krishna2017visual}, gRefCOCO~\cite{liu2023gres},  Open Images V6~\cite{kuznetsova2020open}, \\
                                        & &  V3Det~\cite{wang2023v3det}, \textcolor{blue}{In-house}\\
\addlinespace[4pt]
Classification                              & 2.8M  & \textcolor{blue}{In-house} \\
\bottomrule
\end{tabular}
}

\label{tab:stage1_data}
\end{table}

\begin{table}[h]
\centering
\small
\caption{Details of the training data for multimodal embedding representation learning in stage 2. $^*$ indicates that these datasets all come from MMEB-train~\cite{VLM2Vec}.}
\setlength{\tabcolsep}{2pt}
\resizebox{\linewidth}{!}{
\begin{tabular}{p{3cm}p{2cm}>{\centering\arraybackslash}p{1.8cm}p{10cm}}
\toprule
\textbf{Class} & \textbf{Task} & \textbf{\#Samples} & \textbf{Datasets} \\ 
\midrule

\multirow{2}{*}{\textbf{Single-Modal}} & T$\rightarrow$T (1) & 1M & BAAI-MTP~\cite{baai-mtp} \\
\addlinespace[4pt]
                                       & I$\rightarrow$I (2) & 1.3M & ImageNet-1K~\cite{deng2009imagenet}, NIGHTS$^*$~\cite{fu2023dreamsim} \\

\midrule

\multirow{5}{*}{\textbf{Cross-Modal}}  & \multirow{2}{*}{T$\rightarrow$I (5)}  & \multirow{2}{*}{5.3M} & VisualNews$^*$~\cite{liu2021visual},  MSCOCO$^*$~\cite{lin2014microsoft}, mmE5-synthetic~\cite{chen2025mme5}, VisDial$^*$~\cite{das2017visual},  \\
                                       & & & BLIP Bootstrapped Image-Text Pairs~\cite{li2022blip} \\
                                       \addlinespace[4pt]
                                       & T$\rightarrow$VD (3) & 1.6M & Docmatix~\cite{laurenccon2024building}, Colpali train set~\cite{faysse2024colpali}, VisRAG~\cite{VisRAG} \\
                                       \addlinespace[4pt]
                                       & \multirow{2}{*}{I$\rightarrow$T (7)} & \multirow{2}{*}{0.5M} & ImageNet-1K$^*$~\cite{deng2009imagenet}, HatefulMemes$^*$~\cite{kiela2020hateful}, VOC2007$^*$~\cite{everingham2015pascal}, SUN397$^*$~\cite{xiao2010sun}, \\
                                       & & & VisualNews$^*$~\cite{liu2021visual}, MSCOCO$^*$~\cite{lin2014microsoft}, mmE5-synthetic~\cite{chen2025mme5} \\

\midrule

\multirow{7}{*}{\textbf{Fused-Modal}}  & \multirow{3}{*}{IT$\rightarrow$T (8)} & \multirow{3}{*}{1.6M} & Docmatix~\cite{laurenccon2024building}, mmE5-synthetic~\cite{chen2025mme5}, OK-VQA$^*$~\cite{marino2019ok}, A-OKVQA$^*$~\cite{schwenk2022okvqa}, \\
                                       & & & DocVQA$^*$~\cite{mathew2021docvqa}, InfographicVQA$^*$~\cite{mathew2022infographicvqa}, ChartQA$^*$~\cite{masry2022chartqa}, \\
                                       & & & Visual7W$^*$~\cite{zhu2016visual7w}  \\
                                       \addlinespace[4pt]
                                       & T$\rightarrow$IT (2) & 3.2K & WebQA$^*$~\cite{chang2022webqa}, mmE5-synthetic~\cite{chen2025mme5} \\
                                       \addlinespace[4pt]
                                       & \multirow{2}{*}{IT$\rightarrow$I (5)} & \multirow{2}{*}{5.3M} & MegaPairs~\cite{zhou2025megapairs}, mmE5-synthetic~\cite{chen2025mme5}, CIRR$^*$~\cite{liu2021image}, N24News$^*$~\cite{wang2022n24news}, \\
                                       & & &  MSCOCO$^*$~\cite{lin2014microsoft}  \\
                                       \addlinespace[4pt]
                                       & IT$\rightarrow$IT (1) & 3.1K & mmE5-synthetic~\cite{chen2025mme5} \\
\bottomrule
\end{tabular}
}
\label{tab:stage2_data}
\end{table}
\noindent The training data of stage 2 includes the following three categories. The datasets used for each category are listed in~\cref{tab:stage2_data}.
\begin{itemize}
\item \textbf{Single-Modal:} Includes both Text-to-Text (T$\rightarrow$T) and Image-to-Image (I$\rightarrow$I) pairs.
\item \textbf{Cross-Modal:} The query or candidate is unimodal, and the query–candidate pair spans different modalities, such as Text-to-Image retrieval (T$\rightarrow$I) or Text-to-Visual-Document retrieval (T$\rightarrow$VD).  
\item \textbf{Fused-Modal:} The query and/or candidate contain both image and text. For example, in the MegaPairs dataset~\cite{zhou2025megapairs}, the query is a fusion of an image and a textual instruction, and the target is an image relevant to this mixed image-text query (Image-Text-to-Image, IT$\rightarrow$I).
\end{itemize}

\section{Hyperparameters for Model Implementation}\label{sec:hyper_for_model_impl}

\subsection{Embedder Implementation}\label{subsec:hyper_for_embedder_impl}

In stage 1, we train the full model parameters on 48$\times$NVIDIA A800 (80GB) GPUs with a learning rate of $1\times10^{-5}$ and a global batch size of 48. We set the gradient accumulation steps to 8 and apply dataset packing, training the model for 30,000 steps to restore its generative capability. In stages 2 and 3, we perform contrastive pretraining and task-aware fine-tuning using Low-Rank Adaptation (LoRA) on the same 48$\times$NVIDIA A800 (80GB) GPUs. Both stages use a unified maximum learning rate of $2\times10^{-4}$ and a LoRA rank of 16. Detailed hyperparameters are provided in~\cref{tab:embedding_hyper}. For hard negative judging in stage 3, we employ Qwen3-VL-8B~\cite{li2026qwen3} as the discriminator and insert 12 hard negatives per training instance.

\begin{table}[H]
    \centering
    \setlength\tabcolsep{2pt}
    \caption{Hyperparameters for stage 2 and stage 3 training. ``Warm-Up'' denotes a contrastive warm-up phase in stage 2 using only in-batch negatives; ``Global-HNM'' refers to pretraining in stage 2 with Global Hard Negative Mining; ``MLLM-Judge-FT'' indicates finetuning with an MLLM as a judge.}
    \label{tab:embedding_hyper}
    \resizebox{0.8\linewidth}{!}{
        \begin{tabular}{l>{\centering\arraybackslash}p{1.4cm}>{\centering\arraybackslash}p{1.4cm}>{\centering\arraybackslash}p{1.4cm}>{\centering\arraybackslash}p{1.4cm}>{\centering\arraybackslash}p{1.4cm}>{\centering\arraybackslash}p{1.4cm}}
            \toprule
            \multirow{3}{*}{\textbf{Hyperparameter}} & \multicolumn{2}{c}{\textbf{Stage 2}} & \multicolumn{2}{c}{\textbf{Stage 2}} & \multicolumn{2}{c}{\textbf{Stage 3}} \\
            & \multicolumn{2}{c}{\textbf{(Warm-Up)}} & \multicolumn{2}{c}{\textbf{(Global-HNM)}} & \multicolumn{2}{c}{\textbf{(MLLM-Judge-FT)}} \\
            \cmidrule(lr){2-3} \cmidrule(lr){4-5} \cmidrule(lr){6-7}
            & \textbf{2B} & \textbf{8B} & \textbf{2B} & \textbf{8B} & \textbf{2B} & \textbf{8B} \\
            \midrule
            \#Samples & 16M & 16M & 16M & 16M & 1.5M & 1.5M \\
            \#Hard Negatives & 0 & 0 & 2 & 2 & 12 & 12 \\
            \#GPUs & \multicolumn{6}{c}{48} \\
            Maximum Learning Rate & \multicolumn{6}{c}{$2\times10^{-4}$} \\
            Temperature & \multicolumn{6}{c}{0.03} \\
            LoRA Rank & \multicolumn{6}{c}{16} \\
            Training Epochs & \multicolumn{6}{c}{2} \\
            Batch Size Per Device & 128 & 72 & 64 & 48 & 12 & 10 \\
            \bottomrule
        \end{tabular}}
\end{table}

\subsection{Reranker Implementation}\label{subsec:hyper_for_reranker_impl}

The reranker is initialized from the stage 1 checkpoint. We train it using the same judge-curated data from stage 3 and organize data into pointwise and listwise training. The model is trained with 24$\times$NVIDIA A800 (80GB) GPUs with a learning rate of $4\times10^{-5}$, and the batch size per device is set to 16 and 12 for the 8B and 2B models separately. The model is trained for 2 epochs. The loss weights are set to 1 for both pointwise and listwise objectives. During inference, the reranker is utilized to obtain the two-stage retrieval results by first using the embedder to retrieve a candidate set, followed by a final ranking from the reranker with pointwise reranking of the top-5 results from the embedder.

% ---- Bibliography ----
%
% BibTeX users should specify bibliography style 'splncs04'.
% References will then be sorted and formatted in the correct style.
%
% \bibliographystyle{splncs04}
% \bibliography{main}

\end{document}